\documentclass[11pt]{article}

\usepackage[final]{acl}

\usepackage{times}
\usepackage{latexsym}

\usepackage[T1]{fontenc}

\usepackage[utf8]{inputenc}

\usepackage{microtype}

\usepackage{inconsolata}

\usepackage{graphicx}

\usepackage{amsmath}
\usepackage{amsfonts}
\usepackage{bm}
\usepackage{multirow}
\usepackage{tabularx}
\usepackage{array}
\usepackage{color}
\usepackage{enumitem}
\usepackage{makecell}
\usepackage{xcolor}
\usepackage{colortbl}
\usepackage{booktabs}
\usepackage{algorithm}
\usepackage{algpseudocode}
\usepackage{subcaption}
\usepackage{graphics}
\usepackage{booktabs}
\usepackage{relsize}
\usepackage{xcomment}
\usepackage[most]{tcolorbox}
\usepackage[none]{hyphenat}
\usepackage{etoolbox, soul, xcolor}
\usepackage[dvipsnames]{xcolor}
\usepackage{stfloats}
\usepackage{xurl}

\newcommand{\system}{ConSensus}

\DeclareTextFontCommand{\texttt}{\ttfamily\hyphenchar\font=`\-}

\tcbset{
  modality/.style={
    on line,
    box align=base,
    colframe=#1!65!black,
    colback=#1!15,
    arc=2pt,
    boxsep=0.8pt,
    left=1.6pt,
    right=1.6pt,
    top=1pt,
    bottom=1pt,
    fontupper=\scriptsize\sffamily\bfseries,
    boxrule=0.6pt
  }
}

\newcolumntype{C}{>{\centering\arraybackslash}X}
\newcolumntype{Y}{>{\centering\arraybackslash}p{1.75cm}}
\newcolumntype{L}{>{\raggedright\arraybackslash}l}
\newcolumntype{R}{>{\raggedright\arraybackslash}r}
\definecolor{skyblue}{RGB}{135, 206, 235}

\newbool{showComments}
\booltrue{showComments}

\ifbool{showComments}{%
\newcommand{\hy}[1]{\sethlcolor{lime}\hl{[Jun: #1]}}
\newcommand{\lo}[1]{\sethlcolor{cyan}\hl{[Lorena: #1]}}
\newcommand{\mm}[1]{\sethlcolor{green}\hl{[Mo: #1]}}

}{
\newcommand{\hy}[1]{}
\newcommand{\lo}[1]{}
\newcommand{\mm}[1]{}
}

\newtcolorbox{observationbox}{
  colback=gray!10,
  colframe=gray!50,
  boxrule=0.5pt,
  arc=2pt,
  left=6pt,
  right=6pt,
  top=4pt,
  bottom=4pt
}

\newtcolorbox{promptbox}[1]{
    colback=gray!5,        
    colframe=gray!30,     
    coltitle=black,
    title=#1,             
    arc=1mm,              
    breakable,            
    enhanced,              
    boxsep=4pt,        
    left=4pt,          
    right=4pt,         
    top=4pt,           
    bottom=4pt         
}

\title{\textsl{ConSensus}: Multi-Agent Collaboration for Multimodal Sensing}

\author{
 \textbf{Hyungjun Yoon\textsuperscript{1}}\thanks{Work done during the author's internship at Nokia Bell Labs. Email: hyungjun.yoon@kaist.ac.kr.} \quad
 \textbf{Mohammad Malekzadeh\textsuperscript{2}} \\
 \textbf{Sung-Ju Lee\textsuperscript{1}} \quad
 \textbf{Fahim Kawsar\textsuperscript{2,3}} \quad
 \textbf{Lorena Qendro\textsuperscript{2}} \\ \\
 \textsuperscript{1}KAIST \quad
 \textsuperscript{2}Nokia Bell Labs \quad
 \textsuperscript{3}University of Glasgow \\
}

\begin{document}
\maketitle

\begin{abstract}
\label{sec:abstract}

Large language models (LLMs) are increasingly grounded in sensor data to perceive and reason about human physiology and the physical world. However, accurately interpreting heterogeneous multimodal sensor data remains a fundamental challenge. We show that a single monolithic LLM often fails to reason coherently across modalities, leading to incomplete interpretations and  prior-knowledge bias. We introduce \textbf{\system{}}, a training-free multi-agent collaboration framework that decomposes multimodal sensing tasks into specialized, \textit{modality-aware agents}. To aggregate agent-level interpretations, we propose a \textit{hybrid fusion} mechanism that balances \textit{semantic} aggregation, which enables cross-modal reasoning and contextual understanding, with \textit{statistical} consensus, which provides robustness through agreement across modalities. While each approach has complementary failure modes, their combination enables reliable inference under sensor noise and missing data. We evaluate \system{} on five diverse multimodal sensing benchmarks, demonstrating an average accuracy improvement of 7.1\% over the single-agent baseline. Furthermore, \system{} matches or exceeds the performance of iterative multi-agent debate methods while achieving a 12.7$\times$ reduction in average fusion token cost through a single-round hybrid fusion protocol, yielding a robust and efficient solution for real-world multimodal sensing tasks. The source code is available at \url{https://github.com/nokia/multi-agent-collaboration-for-multimodal-sensing}.

\end{abstract}
\section{Introduction}
\label{sec:introduction}

Large language models~(LLMs) are being extended beyond text to perceive and reason about the physical world and human physiology, motivated by emerging applications in embodied interactions~\citep{embodiedai} and health monitoring~\citep{healthllm}. This requires LLMs to be systematically grounded in sensor data that encodes underlying motions, biosignals, and environmental measurements. Recent studies have explored diverse grounding strategies, including transforming raw signals into descriptive or visual prompts~\citep{fewshothealth, bymyeyes} and jointly aligning learned sensor encoders with LLM representations~\citep{sensorlm}. In contrast to traditional deep learning for sensing~\citep{multimodalsensing}, integration with LLMs enables the resolution of diverse sensing tasks in an interpretable, training-free manner by utilizing the models' internal world knowledge.

\begin{figure}
    \centering
    \includegraphics[width=\columnwidth]{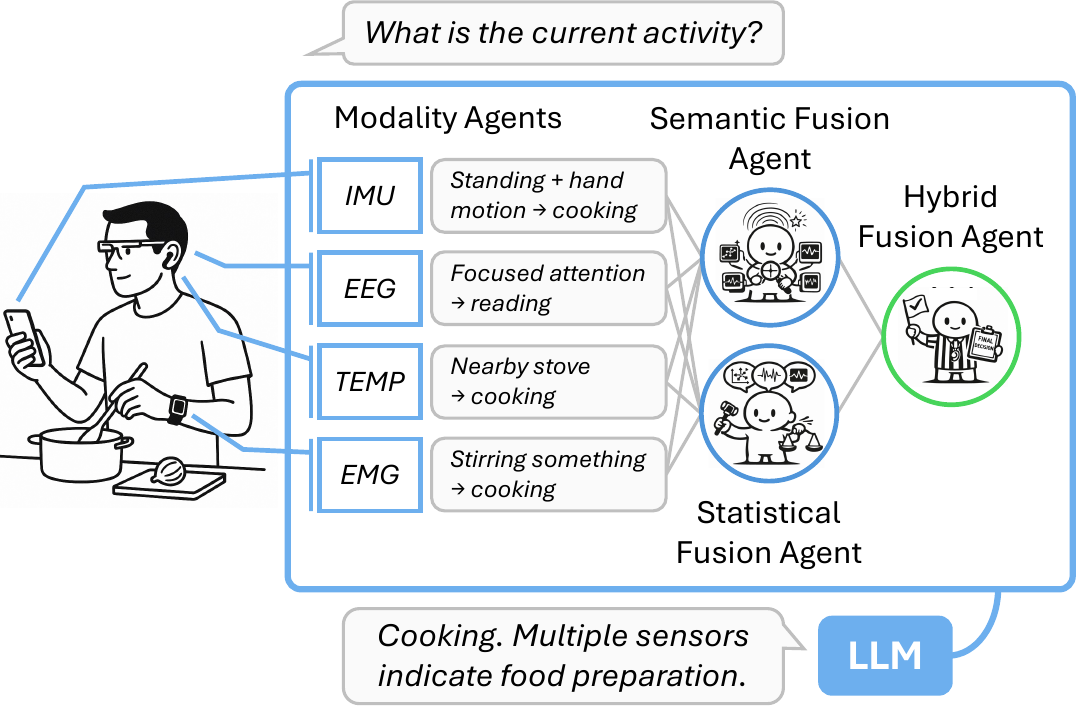}
    \caption{Illustration of \system{}. Modality-aware agents provide specialized interpretations aggregated via hybrid fusion for multimodal sensing.}
    \vspace{-10pt}
    \label{fig:intro_overview}
\end{figure}

Physical and physiological tasks are inherently multimodal, with distinct and complementary aspects captured by heterogeneous sensors~\citep{multimodalsensing}. As a result, multimodal sensing is crucial for combining complementary information across modalities. It becomes particularly important in real-world settings where individual sensors might be unreliable or missing. However, this necessity introduces the challenge of \textit{sensor fusion}: how to systematically integrate heterogeneous modalities into a coherent decision process. Despite growing interest in multimodal LLMs, it remains unclear how LLMs should reason over sensor-specific representations and aggregate their semantic interpretations.

To address the challenge, we leverage multi-agent collaboration as a mechanism for decomposing multimodal reasoning. Recent studies have assigned specialized roles to multiple LLM instances to generate diverse reasoning paths for solving complex tasks~\citep{camel}. However, existing multi-agent frameworks primarily focus on diversifying textual reasoning, and do not address the unique challenges introduced by heterogeneous sensors with asymmetric reliability. In this work, we propose a multi-agent framework for multimodal sensing that explicitly encourages modality-specific agents to produce distinct reasoning trajectories and integrates their semantic interpretations via a structured collaboration protocol.

Our design is guided by three empirically grounded observations. First, we observe that a single LLM struggles to jointly ground its reasoning across all modalities, often producing incomplete interpretations when faced with heterogeneous sensor inputs. This limitation motivates partitioning agents by modality, thereby encouraging independent and complementary reasoning paths. Second, during modality fusion, LLM-based semantic aggregation exhibits pronounced prior-knowledge bias, in which errors from particular modalities are disproportionately amplified and propagated to the final decision. Third, although statistical aggregation mechanisms such as majority voting can mitigate semantic bias, they break down in realistic sensing settings, where missing or unreliable sensor inputs can distort voting outcomes. Taken together, these observations expose a fundamental trade-off between semantic and statistical fusion, motivating a collaboration protocol that balances their complementary strengths and failure modes.

\begin{figure}
    \centering
    \includegraphics[width=0.99\columnwidth]{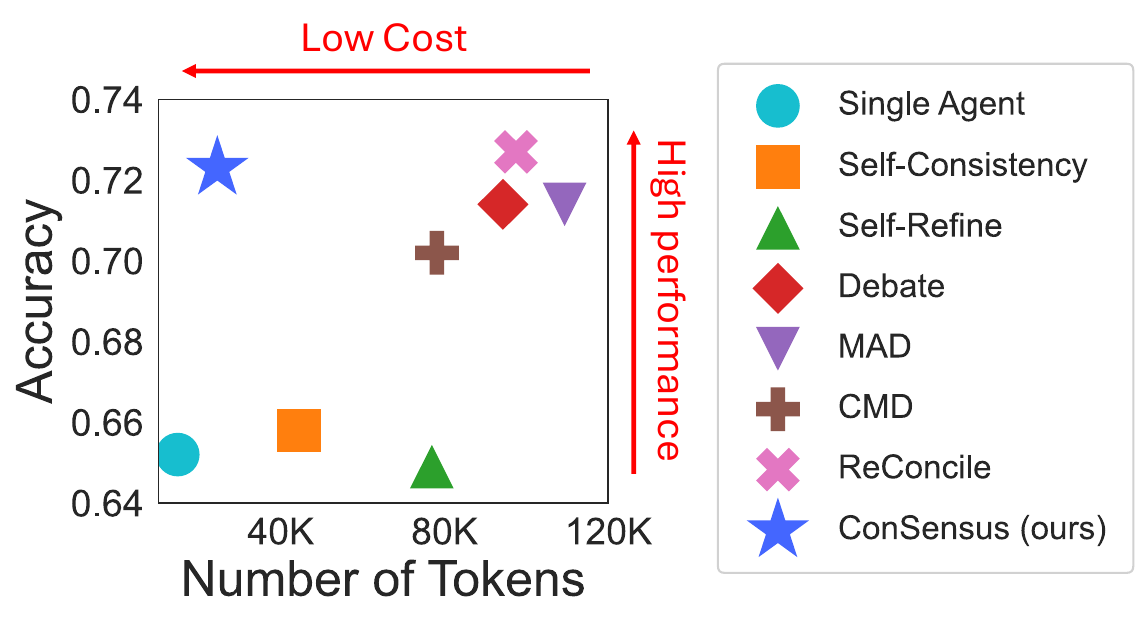}
    \caption{\system{} (top-left) achieves higher accuracy (y-axis) at lower cost (x-axis) compared to baselines.}
    \label{fig:comparison}
\end{figure}

Based on these insights, we present \textbf{\system{}}, a training-free multi-agent framework for multimodal sensing. We define \textit{modality agents} as LLM instances that operate under modality-specific roles and generate independent semantic interpretations of the same task. To aggregate these interpretations into a final decision, we propose a principled \textit{hybrid fusion} strategy. A \textit{semantic fusion agent} performs reasoning-level aggregation to integrate the semantic outputs of modality agents and produce an initial prediction, while a \textit{statistical fusion agent} anchors its decision to a majority-voted outcome across modality agents, providing complementary perspective against prior-driven semantic bias. A final \textit{hybrid fusion agent} is instructed to jointly observe the semantic and statistical fusion outputs to produce the final prediction as a coordinator role.

We evaluate \system{} across five diverse multimodal sensing tasks, covering a broad range of heterogeneous sensor modalities. Experimental results demonstrate that \system{} consistently achieves a $7.1\%$ accuracy improvement over the single-agent baseline. Ablation studies further show that the proposed hybrid fusion mechanism effectively navigates the trade-off between semantic and statistical fusion, adaptively favoring the more reliable reasoning path under varying sensor conditions. Compared to state-of-the-art multi-agent debate baselines applied to the same modality agents, \system{} matches or outperforms accuracy while reducing average fusion token cost by $12.7\times$ (Figure~\ref{fig:comparison}). This efficiency directly follows from our architectural design: while existing multi-agent frameworks depend on costly iterative debates to refine predictions, \system{} employs a single-round, structured fusion protocol, achieving substantially lower inference overhead without sacrificing accuracy.

\begin{figure*}
\centering
\begin{subfigure}[t]{0.33\linewidth}
  \centering
  \includegraphics[width=0.96\linewidth]{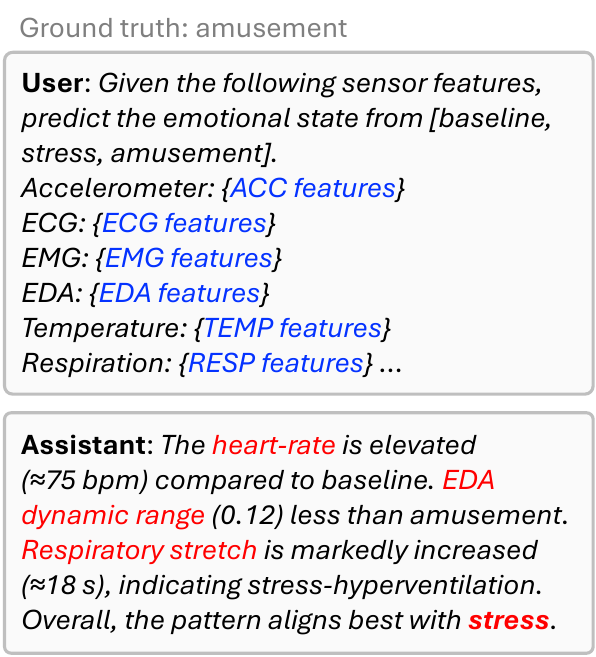}
  \caption{Single-agent inference.}
  \label{fig:motivating_example_1}
\end{subfigure}\hfill
\begin{subfigure}[t]{0.33\linewidth}
  \centering
  \includegraphics[width=0.96\linewidth]{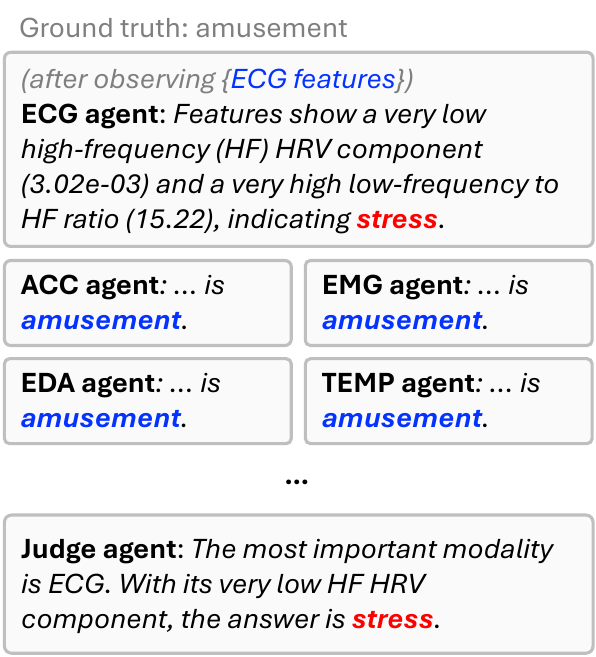}
  \caption{Modality agents with a judge agent.}
  \label{fig:motivating_example_2}
\end{subfigure}\hfill
\begin{subfigure}[t]{0.33\linewidth}
  \centering
  \includegraphics[width=0.96\linewidth]{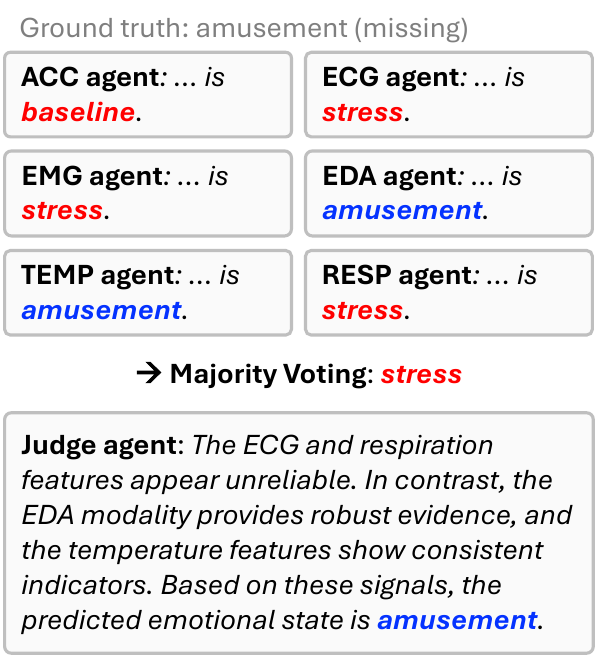}
  \caption{Data with 30\% missing modalities.}
  \label{fig:motivating_example_3}
\end{subfigure}
\vspace{-3pt}
\caption{Examples of LLM-based multimodal sensing on WESAD~\citep{wesad} using \texttt{gpt-oss-20B}.}
\vspace{-8pt}
\label{fig:motivating_examples}
\end{figure*}

Our primary contributions are as follows:
\begin{itemize}[nosep, leftmargin=*]
\item We propose \system{}, the first multi-agent framework for multimodal sensing that decomposes heterogeneous sensor inputs into specialized, modality-aware agents. \system{} operates without additional training, enabling direct deployment across diverse tasks.
\item We present a hybrid fusion strategy that balances the trade-off between semantic aggregation and statistical consensus under sensor uncertainty.
\item We evaluate \system{} across five diverse multimodal sensing tasks, demonstrating an accuracy gain of 7.1\% over the single-agent baseline.
\item We show that our single-round fusion protocol matches or outperforms state-of-the-art multi-agent debate methods while achieving a 12.7$\times$ reduction in fusion token cost, substantially improving inference efficiency and scalability.
\end{itemize}
\section{Background and Motivation}
\label{sec:motivation}

\subsection{Problem Formulation}

We address the \textit{multimodal sensing task}, in which LLMs are leveraged as training-free reasoning engines across diverse tasks to produce semantically interpretable outputs. A multimodal sensing task requires integrating heterogeneous sensor modalities to produce a single, coherent inference.

Formally, given $N$ modalities with inputs $M = \{m_1, m_2, \ldots, m_N\}$ and a task description $T$, the objective is to predict an output $y\in\mathcal{Y}$ that solves the task by jointly reasoning over all available modalities. In this work, each modality input $m_i$ is represented using standard, hand-crafted sensor features (e.g., summary statistics such as \texttt{mean} and \texttt{std}), while noting that our formulation is agnostic to the specific input representation and can be readily extended to alternative forms such as raw signals, images, or learned embeddings~\citep{fewshothealth, bymyeyes, sensorlm}.

\subsection{Motivating Examples}

A straightforward instantiation is to present all modality features ($M$) jointly to a single LLM agent together with the task description $T$. Figure~\ref{fig:motivating_example_1} illustrates an affective state inference task posed to a single agent using multiple sensor modalities~\citep{wesad}. While the agent generates plausible interpretations for some modalities, evidence from other sensors is frequently overlooked, resulting in an incomplete reasoning process and, ultimately, an incorrect prediction. We attribute this failure to \textit{context overload} and \textit{modality dominance} within a single agent, where subtle but critical modality-specific evidence is overridden by salient prominent signals.

\begin{observationbox}
    \noindent \textbf{Observation 1:} A single agent often produces an \textit{incomplete} cross-modal interpretation in multimodal sensing tasks.
\end{observationbox}

To ensure completeness, we decompose multimodal interpretation into \textit{modality agents}, each of which follows an independent reasoning path conditioned on a specific sensor modality ($m_i$). This decomposition produces multiple semantic interpretations for the same task, which must subsequently be aggregated into a single decision. Common aggregation strategies include (i)~employing an additional \textit{judge agent}~\citep{mad} to synthesize a final decision from all agent outputs, or (ii)~applying majority voting~\citep{debate} over individual predictions.

\begin{figure*}
    \centering
    \includegraphics[width=0.95\linewidth]{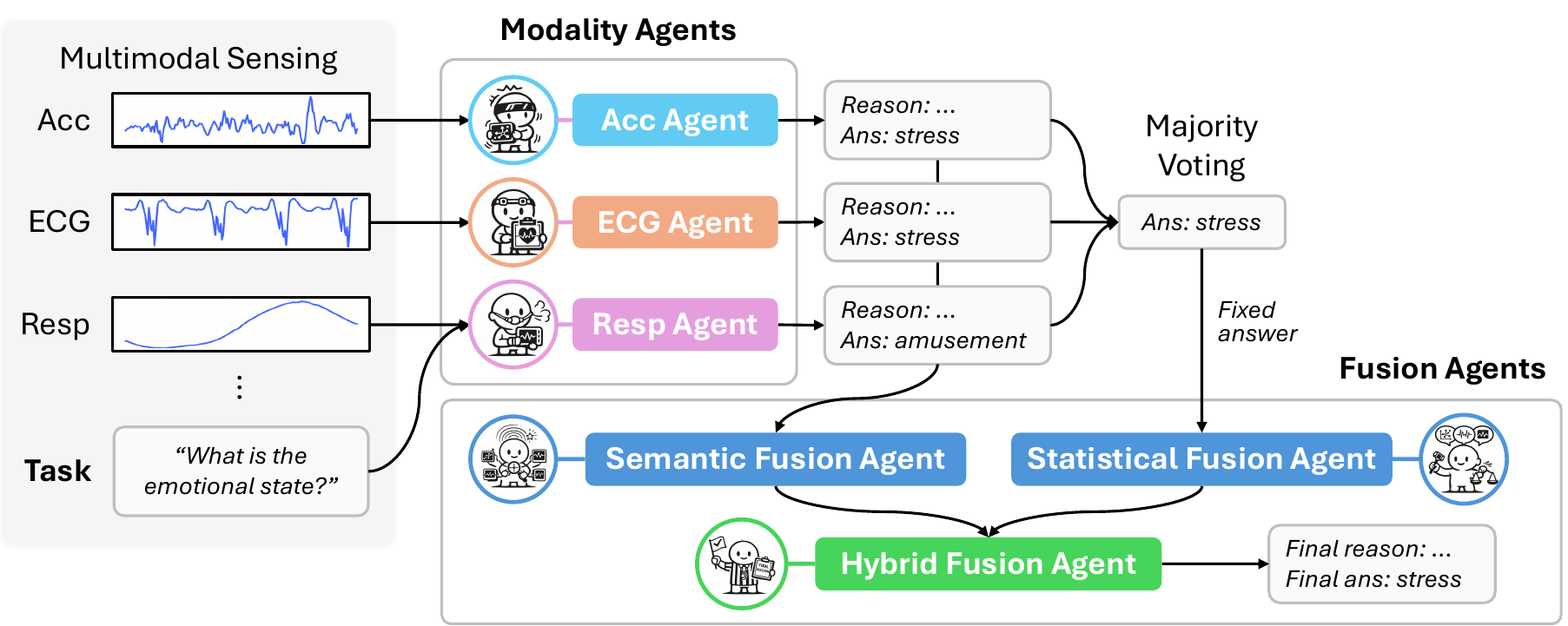}
    \vspace{-3pt}
    \caption{Overview of \system{}: (i) Modality agents generate specialized per-sensor interpretations; (ii) a semantic fusion agent aggregates cross-modal reasoning; (iii) a statistical fusion agent provides an output that anchors the reasoning to the majority; and (iv) a hybrid fusion agent reconciles both outputs to yield the final decision.}
    \vspace{-9pt}
    \label{fig:overview}
\end{figure*}

However, we find that the judge agent often relies on prior domain knowledge to disproportionately weight certain modalities (e.g., privileging clinically salient signals such as ECG), leading to incorrect decisions when those modalities are unreliable (Figure~\ref{fig:motivating_example_2}). In real-world deployments, sensor signals often deviate from canonical patterns due to device-specific characteristics or environmental noise. While complementary modalities with consistent evidence can mitigate such noise, the LLM's \textit{knowledge bias} frequently causes error propagation; incorrect assumptions associated with a perceived ``important'' modality override correct evidence from other modalities.

\vspace{-5pt}
\begin{observationbox}
    \noindent \textbf{Observation 2:} An LLM-based judge is prone to \textit{knowledge bias}, whereby prior domain knowledge can mislead consensus even when robust evidence exists in other modalities.
\end{observationbox}
\vspace{-5pt}

A natural approach to mitigating knowledge bias is majority voting, which relies on statistical aggregation rather than semantic reasoning or prior knowledge. Majority voting is a simple and widely used form of late fusion that can be effective when multiple modalities provide reliable and complementary evidence. From this perspective, aggregating modality-level predictions through agreement offers a practical way to improve robustness against individual reasoning errors.

Conversely, when modality errors are correlated or sensor inputs are corrupted or missing (i.e., \textit{sensor failure}), which is common in real-world sensing~\citep{missing_data}, majority voting can fail catastrophically. Figure~\ref{fig:motivating_example_3} illustrates such a failure case under 30\% missing modalities. Sensor failures violate the core assumption of majority voting~\citep{cjt} (i.e., voter reliability and independence), causing statistical aggregation to converge toward an incorrect consensus. In contrast, the judge agent identifies uncertainty cues to discount unreliable reasoning and prioritize complementary modalities that provide consistent rationales.

\begin{observationbox}
    \noindent \textbf{Observation 3:} Majority voting is prone to \textit{sensor failure}, often producing unreliable votes and false consensus in multimodal sensing.
\end{observationbox}

Together, observations~1--3 reveal complementary failure modes in multimodal reasoning: single-agent reasoning suffers from incomplete cross-modal interpretation (Observation~1), semantic aggregation is vulnerable to knowledge bias (Observation~2), and statistical aggregation is fragile under sensor failures (Observation~3). These observations jointly expose a fundamental tension between semantic-level and statistical aggregation.

\section{Method}
\label{sec:method}

We introduce \textbf{\system{}}, a training-free \underline{Con}versational \underline{Sens}or F\underline{us}ion framework that performs multimodal sensing via structured, role-specialized multi-agent collaboration (Figure~\ref{fig:overview}). \system{} comprises four defined agent roles: (i)~\textbf{\textit{modality agents}}, each specialized to a single sensor modality and responsible for producing task-relevant semantic interpretations; (ii)~a \textbf{\textit{semantic fusion agent}} that integrates modality-agent outputs into a holistic prediction by reasoning over cross-modal semantic evidence; (iii)~a \textbf{\textit{statistical fusion agent}} that reasons over a fixed prediction anchored to a majority-vote consensus of modality-level predictions, providing robustness against prior-driven knowledge bias; and (iv)~a \textbf{\textit{hybrid fusion agent}} that arbitrates between the outputs of the semantic and statistical fusion agents. By explicitly balancing knowledge-grounded and consensus-grounded reasonings, the hybrid fusion agent produces the final prediction. 

\subsection{Modality Agents}
\label{sec:method_modality_agents}

\system{} instantiates a set of \textbf{modality agents} $\{A^{\text{mod}}_1, A^{\text{mod}}_2, \ldots, A^{\text{mod}}_N\}$, where each agent $A^{\text{mod}}_i$ is assigned to a single sensor modality $m_i$ and is responsible for producing a task-relevant semantic interpretation of that modality under the query $T$. Each agent receives a modality-conditioned prompt $(m_i, T)$ and outputs (i)~a modality-specific prediction $\hat{y}_i \in \mathcal{Y}$ and (ii)~a rationale $r_i$ that explicitly grounds the prediction in evidence. This modularization isolates modality-level reasoning, reducing cross-modal interference within the prompt and preserving complementary evidence that may otherwise be diluted under joint prompting.

\subsection{Fusion Agents}
\label{sec:method_fusion_agents}

To reach a final prediction, \system{} aggregates the outputs of the modality agents, $\{(\hat{y}_i, r_i)\}_{i=1}^N$. As discussed in Section~\ref{sec:motivation}, semantic and statistical aggregation paradigms exhibit complementary strengths and distinct failure modes. To balance these trade-offs, we define a \emph{hybrid fusion} approach comprising three fusion agents.

\noindent \textbf{Semantic Fusion Agent ($A^{\text{fuse}}_{\text{sem}}$).} It implements a \textit{knowledge-grounded aggregation} that synthesizes semantic evidence across modality-agent outputs 
$\{(\hat{y}_1, r_1), \ldots, (\hat{y}_N, r_N)\}$ using the LLM's prior knowledge.
It represents a semantic inductive bias that emphasizes cross-modal coherence, causal plausibility, and high-level domain knowledge when forming an aggregated hypothesis.

\noindent  \textbf{Statistical Fusion Agent ($A^{\text{fuse}}_{\text{stat}}$).} It implements a \textit{consensus-grounded aggregation} by anchoring its reasoning to a fixed majority-voted prediction $\hat{y}_{\text{vote}} = \arg\max_{c \in \mathcal{Y}} \sum_{i=1}^{N} \mathbf{1}[\hat{y}_i = c]$. Rather than generating a new prediction, it produces a rationale that justifies this statistical consensus, representing a complementary inductive bias that prioritizes robustness to individual agent errors and suppresses prior-driven knowledge bias. In our framework, majority voting is used as a simple representative form of agreement-based aggregation.

\noindent  \textbf{Hybrid Fusion Agent ($A^{\text{fuse}}_{\text{hyb}}$).} We formalize hybrid fusion as an instance-wise arbitration between two complementary inductive biases: $(\hat{y}_{\text{sem}}, r_{\text{sem}}) = A^{\text{fuse}}_{\text{sem}}(\{(\hat{y}_i, r_i)\}_{i=1}^N)$ and $(\hat{y}_{\text{stat}}, r_{\text{stat}}) = A^{\text{fuse}}_{\text{stat}}(\{(\hat{y}_i, r_i)\}_{i=1}^N)$. The hybrid fusion agent $A^{\text{fuse}}_{\text{hyb}}$ performs instance-wise arbitration: $(\hat{y}, r) = A^{\text{fuse}}_{\text{hyb}}\big((\hat{y}_{\text{sem}}, r_{\text{sem}}), (\hat{y}_{\text{stat}}, r_{\text{stat}})\big)$ which selects between two complementary aggregation hypotheses grounded in (i) semantic reasoning and (ii) statistical consensus as a \textit{coordinator} role.

Unlike prior methods that rely only on knowledge-grounded reasoning, the hybrid fusion agent is additionally exposed to statistically grounded reasoning. This design introduces a complementary, data-driven perspective into the decision process, allowing \system{} to maintain robustness under sensor failure via semantic coherence, while simultaneously mitigating knowledge bias by taking account of statistical stability. 

\section{Experiments}
\label{sec:experiments}

\subsection{Setup}
\label{sec:exp_setups}

\noindent \textbf{Datasets.} We evaluate \system{} across five diverse multimodal sensing benchmarks, spanning both physiological and physical activity recognition tasks: (i)~WESAD~\citep{wesad} for affective state recognition; (ii)~SleepEDF~\citep{sleepedf} for sleep stage classification; (iii)~ActionSense~\citep{actionsense} for recognizing four categories of kitchen activities (e.g., peeling, washing); (iv)~MMFit~\citep{mmfit} for gym exercise recognition; and (v)~PAMAP2~\citep{pamap2} for daily activity recognition. The datasets cover a wide range of 12 distinct sensor modalities collected from different locations and devices (e.g., smartphones, smartwatches, and earbuds). Raw sensor data were processed into standard, hand-crafted feature representations following established protocols~\citep{wesad, sleepedf_features}. We detail the sensor modalities, preprocessing, and splits in Appendix~\ref{sec:appendix_dataset}.

\noindent \textbf{Prompts.} We employ text-only prompts that embed extracted sensor features within a structured prompt template. Each prompt includes a detailed description of the target task and data, derived directly from the original dataset documentation. 
Furthermore, we employ a 1-shot in-context learning strategy, providing one representative example per class to enhance the model reasoning. We provide the prompt details in Appendix~\ref{sec:appendix_prompts}.

\noindent \textbf{Models and Metrics.} We use \texttt{gpt-oss-20B} as the main backbone~\citep{gptoss}. Moreover, we evaluate scale and architecture variations using \texttt{gpt-oss-120B}, \texttt{Llama-3.1-8B-Instruct}~\citep{llama3} and \texttt{Llama-4-Scout-17B-16E}~\citep{llama4}. We use a temperature of 0 to ensure deterministic and reproducible outputs. As the datasets are class-balanced, we use accuracy as the evaluation metric. We report macro-F1 in Appendix~\ref{sec:appendix_f1score}, which shows trends consistent with accuracy.

\noindent \textbf{Baselines.} We evaluate \system{} against three representative baselines: (i)~\textbf{Single-Agent}, where all modality features are concatenated into a single prompt, (ii)~\textbf{Self-Consistency} (SC)~\citep{sc}, which samples multiple reasoning paths and selects the most consistent prediction (with temp$=0.7$), and (iii)~\textbf{Self-Refine} (SR)~\citep{sr}, which iteratively generates feedback and refines an initial response. For SC, we use three LLM instances, and for SR, we perform two refinement steps to ensure a comparable inference budget with \system{}, which involves two intermediate fusion steps prior to the final hybrid decision.

\newcolumntype{D}{>{\centering\arraybackslash}m{1.8cm}}
\begin{table*}
\centering
\renewcommand{\arraystretch}{1.0}
\smaller
\begin{tabularx}{0.96\textwidth}{LD@{\hspace{6pt}}D@{\hspace{6pt}}D@{\hspace{6pt}}D@{\hspace{6pt}}D@{\hspace{6pt}}D}
\toprule
Method & WESAD & SleepEDF & ActionSense & MMFit & PAMAP2 & Avg. \\
\midrule

\rowcolor{gray!10}
\multicolumn{7}{l}{\textit{Single-agent} baselines} \\
\rowcolor{white}
Single-Agent & 0.793 \scalebox{0.8}{$\pm$ 0.033} & 0.519 \scalebox{0.8}{$\pm$ 0.031} & 0.577 \scalebox{0.8}{$\pm$ 0.032} & 0.819 \scalebox{0.8}{$\pm$ 0.022} & 0.551 \scalebox{0.8}{$\pm$ 0.027} & 0.652 \scalebox{0.8}{$\pm$ 0.027} \\
$+$ Self-Consistency & 0.786 \scalebox{0.8}{$\pm$ 0.035} & 0.541 \scalebox{0.8}{$\pm$ 0.031} & 0.555 \scalebox{0.8}{$\pm$ 0.031} & 0.862 \scalebox{0.8}{$\pm$ 0.019} & 0.547 \scalebox{0.8}{$\pm$ 0.027} & 0.658 \scalebox{0.8}{$\pm$ 0.027} \\
$+$ Self-Refine & 0.747 \scalebox{0.8}{$\pm$ 0.035} & 0.551 \scalebox{0.8}{$\pm$ 0.031} & 0.566 \scalebox{0.8}{$\pm$ 0.031} & 0.822 \scalebox{0.8}{$\pm$ 0.022} & 0.563 \scalebox{0.8}{$\pm$ 0.026} & 0.650 \scalebox{0.8}{$\pm$ 0.026} \\

\midrule

\rowcolor{gray!10}
\multicolumn{7}{l}{\textit{Modality agents} $+$ multi-agent debate baselines, requiring \textcolor{red}{\textbf{76K additional tokens}} per sample} \\
\rowcolor{red!8}
$+$ Debate & 0.873 \scalebox{0.8}{$\pm$ 0.027} & 0.548 \scalebox{0.8}{$\pm$ 0.032} & 0.609 \scalebox{0.8}{$\pm$ 0.031} & \textbf{0.984} \scalebox{0.8}{$\pm$ 0.007} & 0.561 \scalebox{0.8}{$\pm$ 0.028} & 0.715 \scalebox{0.8}{$\pm$ 0.028} \\
\rowcolor{red!8}
$+$ MAD & 0.847 \scalebox{0.8}{$\pm$ 0.029} & 0.562 \scalebox{0.8}{$\pm$ 0.031} & \underline{0.613} \scalebox{0.8}{$\pm$ 0.031} & 0.960 \scalebox{0.8}{$\pm$ 0.011} & \textbf{0.589} \scalebox{0.8}{$\pm$ 0.026} & 0.714 \scalebox{0.8}{$\pm$ 0.026} \\
\rowcolor{red!8}
$+$ CMD & 0.840 \scalebox{0.8}{$\pm$ 0.030} & 0.578 \scalebox{0.8}{$\pm$ 0.031} & 0.589 \scalebox{0.8}{$\pm$ 0.034} & 0.962 \scalebox{0.8}{$\pm$ 0.011} & 0.539 \scalebox{0.8}{$\pm$ 0.025} & 0.702 \scalebox{0.8}{$\pm$ 0.025} \\
\rowcolor{red!8}
$+$ ReConcile & \underline{0.880} \scalebox{0.8}{$\pm$ 0.027} & 0.571 \scalebox{0.8}{$\pm$ 0.030} & \textbf{0.640} \scalebox{0.8}{$\pm$ 0.031} & 0.964 \scalebox{0.8}{$\pm$ 0.011} & \underline{0.579} \scalebox{0.8}{$\pm$ 0.026} & \textbf{0.727} \scalebox{0.8}{$\pm$ 0.026} \\

\midrule

\rowcolor{gray!10}
\multicolumn{7}{l}{\textit{Modality agents} $+$ fusion agents (ours), requiring \textcolor{blue}{\textbf{6K additional tokens}} per sample} \\
\rowcolor{blue!8}
$+$ Semantic Fusion & 0.825 \scalebox{0.8}{$\pm$ 0.031} & 0.580 \scalebox{0.8}{$\pm$ 0.031} & 0.605 \scalebox{0.8}{$\pm$ 0.033} & 0.964 \scalebox{0.8}{$\pm$ 0.011} & 0.559 \scalebox{0.8}{$\pm$ 0.026} & 0.707 \scalebox{0.8}{$\pm$ 0.026} \\
\rowcolor{blue!8}
$+$ Statistical Fusion & \textbf{0.927} \scalebox{0.8}{$\pm$ 0.021} & \underline{0.592} \scalebox{0.8}{$\pm$ 0.032} & 0.597 \scalebox{0.8}{$\pm$ 0.033} & 0.960 \scalebox{0.8}{$\pm$ 0.011} & 0.534 \scalebox{0.8}{$\pm$ 0.026} & 0.722 \scalebox{0.8}{$\pm$ 0.026} \\
\rowcolor{blue!8}
\textbf{\system{}} & \underline{0.880} \scalebox{0.8}{$\pm$ 0.029} & \textbf{0.600} \scalebox{0.8}{$\pm$ 0.031} & 0.611 \scalebox{0.8}{$\pm$ 0.031} & \underline{0.967} \scalebox{0.8}{$\pm$ 0.010} & 0.558 \scalebox{0.8}{$\pm$ 0.026} & \underline{0.723} \scalebox{0.8}{$\pm$ 0.026} \\

\bottomrule
\end{tabularx}
\caption{Comparison of \system{} with baseline methods. Ablation results for the individual semantic and statistical fusion agents are reported for comparison. The best and second-best accuracies are \textbf{bold} and \underline{underlined}.}
\label{tab:main_results}
\end{table*}

To further evaluate the effectiveness of our hybrid fusion design, we compare \system{} against state-of-the-art multi-agent debate frameworks adapted to operate over the same set of modality agents: (i)~\textbf{Debate}~\citep{debate}, where agents iteratively refine responses by observing the outputs of other agents, with the final decision determined by majority voting; (ii)~\textbf{MAD}~\citep{mad}, which follows a similar protocol to Debate but employs a judge agent to steer the final decision; (iii)~\textbf{CMD}~\citep{cmd}, where agents are partitioned into groups (we use two groups) such that agents within each group share full responses, while only prediction counts are exchanged across groups; and (iv)~\textbf{ReConcile}~\citep{reconcile}, where agents output an explicit confidence score and arrive at a final decision via confidence-weighted voting. As SC and SR, we fix the number of debate rounds to two. We emphasize that, while these baselines are designed for multi-round debates, \system{} performs aggregation in a single round, highlighting a fundamental difference in both efficiency and interaction structure.

Importantly, \system{} is a \textit{training-free} framework. By leveraging the world knowledge and reasoning capabilities of pre-trained LLMs, our method bypasses the need for large-scale data collection and task-specific model training. Consequently, we do not include traditional learning-based sensor fusion methods~\citep{multimodalsensing} as direct baselines, since they rely on supervised training over substantial task-specific data. Nevertheless, to contextualize performance against a supervised reference, we provide a comparison in Appendix~\ref{app:supervised_fusion}.

\subsection{Results}
\label{sec:exp_results}

\noindent \textbf{Effect of Modality Agents.} We first evaluate the impact of modality agents by comparing \system{} against single-agent reasoning baselines. As shown in Table~\ref{tab:main_results}, \system{} consistently outperforms the Single-Agent baseline by an average accuracy margin of $7.1\%$. Notably, even without hybrid fusion, both semantic-only and statistical-only fusion variants achieve substantial performance gains, indicating that modality-specific agent decomposition alone yields strong improvements. While Self-Consistency yields only marginal improvements (an average gain of $0.6\%$), its performance remains substantially below that of \system{}. Qualitative analysis (Appendix~\ref{sec:appendix_outputs}) shows that the Single-Agent baseline often omits modality-specific interpretations, consistent with Observation~1. In contrast, \system{} ensures complete modality coverage, resulting in more reliable and well-grounded decisions.

\noindent \textbf{Effect of Hybrid Fusion.} Table~\ref{tab:main_results} further compares \system{} with semantic-only and statistical-only fusion variants. While statistical fusion outperforms semantic fusion by an average margin of 1.5\%, the optimal fusion strategy varies substantially across datasets. Semantic fusion performs best when sensor data align well with LLM prior knowledge (e.g., PAMAP2, $2.5\%$), whereas statistical fusion dominates under unexpected or noisy data characteristics (e.g., WESAD, $10.2\%$). This is consistent with Observation~2: in WESAD, the semantic fusion agent frequently overweights incorrect ECG-derived predictions due to knowledge bias, whereas majority voting aggregates correct evidence from secondary modalities.

Importantly, the hybrid fusion agent resolves this trade-off, achieving the highest average accuracy across datasets. On SleepEDF, ActionSense, and MMFit, the hybrid agent surpasses both semantic-only and statistical-only fusion agents. This demonstrates the hybrid agent's ability to follow semantic reasoning when it is reliable and to revert to statistical consensus when knowledge bias would otherwise lead to incorrect decisions.

\begin{figure}[t]
    \centering
    \includegraphics[width=0.9\columnwidth]{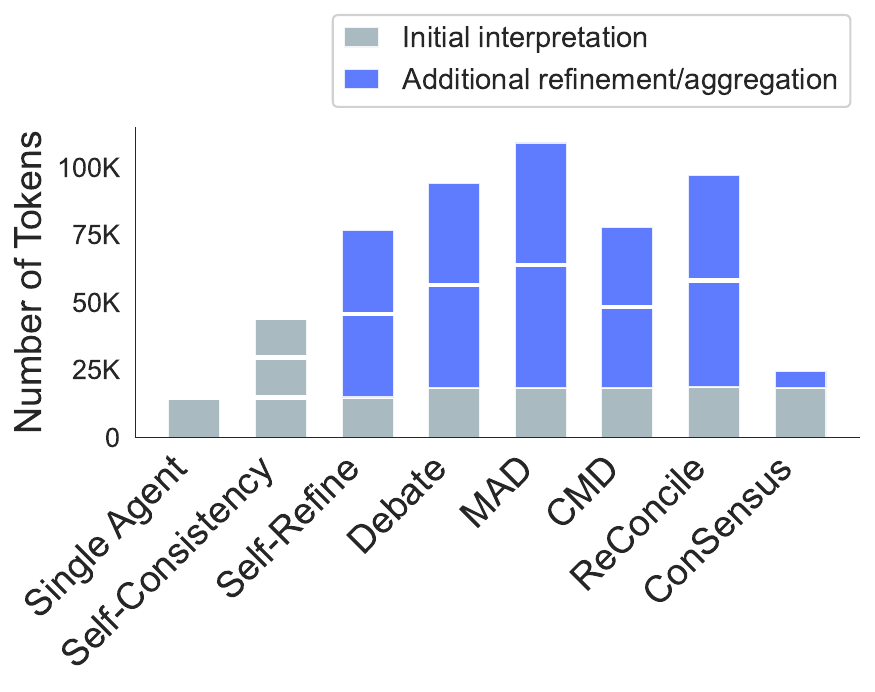}
    \caption{Average input tokens per inference across datasets. Gray bars denote tokens required for initial interpretation, and blue bars denote aggregation or refinement tokens, segmented by rounds.}
    \label{fig:token_efficiency}
\end{figure}

\noindent \textbf{Comparison with Multi-Agent Debate.} We further compare \system{} with iterative multi-agent debate baselines built on the same set of modality agents. \system{} achieves the highest accuracy among all baselines, with the exception of ReConcile, which attains comparable performance ($72.7\%$ vs. $72.3\%$). Notably, all debate baselines rely on multi-round interactions, incurring substantial token overhead. In contrast, \system{} achieves comparable or superior accuracy using a single-round fusion protocol, resulting in up to $12.7\times$ reduction in fusion tokens compared to the debate baselines. We further evaluate non-iterative variants of the debate baselines 
in Appendix~\ref{sec:appendix_round0}, showing that \system{} consistently outperforms debate baselines under similar number of tokens.

\begin{figure}
    \centering
    \includegraphics[width=0.9\columnwidth]{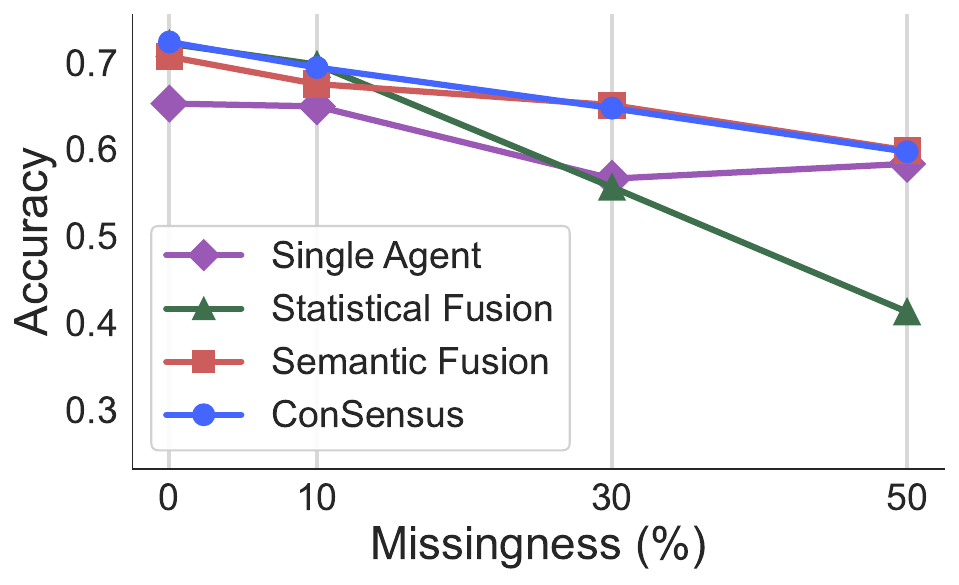}
    \caption{Accuracy under random modality omission at varying missingness levels. \system{} maintains higher accuracy by adaptively balancing semantic and statistical fusion.}
    \label{fig:missingness_comparison}
\end{figure}

\noindent \textbf{Token Efficiency.} Figure~\ref{fig:token_efficiency} reports the average token usage per inference across all datasets. Gray bars represent tokens consumed during initial interpretation, including system prompts and sensor feature descriptions required to derive an initial prediction. These costs remain consistent across all methods, as modality agents partition the same underlying sensory information rather than introducing additional content. Blue bars represent tokens consumed by refinement, debate, or fusion procedures. 
While ReConcile, one of the strongest baselines, requires 78.6K tokens for aggregation per inference to achieve accuracy comparable to \system{}, \system{} requires only 6K tokens. 
Averaged across all multi-agent debate baselines, \system{} reduces aggregation tokens by $12.7\times$. This result highlights the substantial token efficiency of \system{}, achieved through its single-round structured fusion design.

\noindent \textbf{Robustness to Sensor Failure.} We simulate sensor failure by randomly omitting 10\%, 30\%, and 50\% of sensor modalities. As shown in Figure~\ref{fig:missingness_comparison}, \system{} maintains a consistent performance advantage, outperforming the Single-Agent baseline by an average of $7.1\%$, $4.5\%$, $8.2\%$, $1.2\%$ across the respective missingness ratios. 

We observe that the statistical fusion degrades sharply as sensor missingness increases, collapsing to 41.4\% at 50\% missingness. This confirms Observation~3: majority voting is fragile when anchored to a high ratio of unreliable votes. In contrast, semantic fusion exhibits remarkable resilience, maintaining 59.9\% accuracy even in extreme failure scenarios. Importantly, \system{}'s hybrid fusion dynamically prioritizes the semantic reasoning as statistical certainty drops. Consequently, \system{} outperforms the statistical fusion by 9.1\% at 30\% and 18.4\% at 50\% missingness, effectively mitigating catastrophic degradation by selectively following the semantic fusion agent.

We additionally examine robustness to sensor corruption in Appendix~\ref{app:noise_robustness} by evaluating \system{} under additive white Gaussian noise (AWGN) at 10 dB SNR. The results show broadly similar performance trends to those in the main setting, indicating that \system{} is robust to both modality missingness and moderate signal perturbation.

\newcolumntype{S}{>{\centering\arraybackslash}m{0.6cm}}
\begin{table}[t]
\centering
\renewcommand{\arraystretch}{1.0}
\scriptsize
\begin{tabularx}{0.95\columnwidth}{L@{\hspace{6pt}}S@{\hspace{6pt}}S@{\hspace{6pt}}S@{\hspace{6pt}}S@{\hspace{6pt}}S@{\hspace{6pt}}S}
\toprule
Method & \shortstack{WE-\\SAD} & \shortstack{Sleep\\EDF} & \shortstack{Action\\Sense} & \shortstack{MM-\\Fit} & \shortstack{PA-\\MAP2} & Avg. \\
\midrule


\multicolumn{7}{l}{\raisebox{-0.2\height}{\includegraphics[height=8pt]{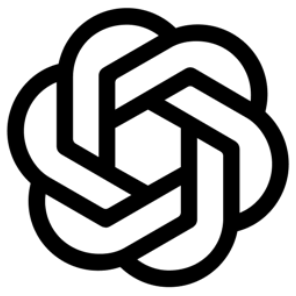}} \texttt{gpt-oss-120B}} \\
\addlinespace[2pt]
\rowcolor{gray!10}
\multicolumn{7}{l}{\textit{Single agent}} \\
Single-Agent & 0.807 & 0.580 & 0.564 & 0.730 & 0.592 & 0.654 \\
\rowcolor{gray!10}
\multicolumn{7}{l}{\textit{Modality agents}} \\
\rowcolor{red!8}
$+$ ReConcile & \textbf{0.887} & \textbf{0.588} & \textbf{0.604} & \textbf{0.980} & \textbf{0.636} & \textbf{0.739} \\
\rowcolor{blue!8}
$+$ Semantic Fusion & 0.793 & 0.572 & 0.600 & 0.967 & 0.633 & 0.713 \\
\rowcolor{blue!8}
$+$ Statistical Fusion & 0.880 & 0.580 & 0.572 & 0.960 & 0.572 & 0.713 \\
\rowcolor{blue!8}
\textbf{\system{}} & 0.833 & 0.572 & 0.584 & 0.967 & 0.611 & 0.713 \\
\midrule

\multicolumn{7}{l}{\raisebox{-0.2\height}{\includegraphics[height=8pt]{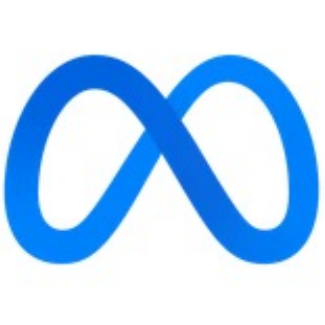}} \texttt{Llama-3.1-8B-Instruct}} \\
\addlinespace[2pt]
\rowcolor{gray!10}
\multicolumn{7}{l}{\textit{Single agent}} \\
Single-Agent & 0.493 & 0.292 & 0.288 & 0.167 & 0.225 & 0.293 \\
\rowcolor{gray!10}
\multicolumn{7}{l}{\textit{Modality agents}} \\
\rowcolor{red!8}
$+$ ReConcile & 0.673 & 0.304 & 0.296 & 0.390 & 0.231 & 0.379 \\
\rowcolor{blue!8}
$+$ Semantic Fusion & 0.773 & 0.368 & 0.288 & 0.493 & 0.344 & 0.453 \\
\rowcolor{blue!8}
$+$ Statistical Fusion & \textbf{0.787} & \textbf{0.416} & \textbf{0.324} & \textbf{0.543} & 0.342 & \textbf{0.482} \\
\rowcolor{blue!8}
\textbf{\system{}} & 0.780 & 0.368 & 0.296 & 0.487 & \textbf{0.347} & 0.456 \\
\midrule

\multicolumn{7}{l}{\raisebox{-0.2\height}{\includegraphics[height=8pt]{figures/llama_icon.pdf}} \texttt{Llama-4-Scout-17B-16E}} \\
\addlinespace[2pt]
\rowcolor{gray!10}
\multicolumn{7}{l}{\textit{Single agent}} \\
Single-Agent & 0.660 & 0.416 & 0.268 & 0.293 & 0.386 & 0.405 \\
\rowcolor{gray!10}
\multicolumn{7}{l}{\textit{Modality agents}} \\
\rowcolor{red!8}
$+$ ReConcile & 0.853 & 0.556 & 0.572 & 0.877 & 0.497 & 0.671 \\
\rowcolor{blue!8}
$+$ Semantic Fusion & 0.833 & 0.536 & 0.588 & 0.900 & 0.525 & 0.676 \\
\rowcolor{blue!8}
$+$ Statistical Fusion & \textbf{0.860} & \textbf{0.568} & \textbf{0.604} & \textbf{0.910} & 0.531 & \textbf{0.695} \\
\rowcolor{blue!8}
\textbf{\system{}} & \textbf{0.860} & 0.548 & 0.592 & 0.903 & \textbf{0.542} & 0.689 \\

\bottomrule
\end{tabularx}
\vspace{-3pt}
\caption{Performance comparison of Single-Agent baseline, ReConcile, Semantic Fusion, Statistical Fusion, and \system{} across four LLM backbones.}
\vspace{-12pt}
\label{tab:llm_comparison}
\end{table}

\noindent \textbf{Effect of LLM.} We evaluate \system{} across diverse open-sourced LLMs, including 
\texttt{gpt-oss-120B}, \texttt{Llama-3.1-8B-Instruct}, and the Mixture-of-Experts (MoE) based \texttt{Llama-4-Scout-17B-16E} (109B total parameters). Table~\ref{tab:llm_comparison} compares \system{} against the Single-Agent baseline, the best-performing debate baseline ReConcile, and the semantic and statistical fusion agents.

Across all LLMs, \system{} consistently outperforms the Single-Agent baseline and exhibits robust hybrid fusion behavior, effectively balancing semantic and statistical fusion results. For \texttt{gpt-oss-120B}, semantic and statistical fusion achieve comparable performance, and ReConcile outperforms \system{}, suggesting that the prior knowledge in larger parameters benefit the fusion and further debates. In contrast, on \texttt{Llama} models, \system{} consistently outperforms ReConcile while requiring fewer tokens (up to $13.1\times$), indicating that ReConcile's advantages are not robust across LLMs. Notably, on the small \texttt{Llama-3.1-8B-Instruct} model, the Single-Agent baseline performs poorly (29.3\%), and ReConcile yields limited improvement (+8.6\%), whereas \system{} achieves a substantially larger gain (+16.3\%). This highlights the strong potential of \system{} to enhance multimodal sensing capabilities for smaller, resource-constrained LLMs. Overall, \system{} delivers consistent performance gains across model families while maintaining high token efficiency.

\section{Related Work}
\label{sec:related_work}

\subsection{LLM-based Sensing}

Grounding LLMs with sensor data enables a broad spectrum of sensing tasks by leveraging world knowledge and semantic reasoning to interpret rich real-world contexts. Early works~\citep{penetrativeai, fewshothealth} demonstrated that LLMs can interpret raw or processed sensor data to solve real-world sensing problems. Subsequent studies have explored broader reasoning capabilities over sensor inputs~\citep{llmsense} , conversational analysis over long-term sensor traces~\citep{sensorchat}. To improve sensor understanding, recent works proposed transforming sensor data into visual representations~\citep{bymyeyes}, aligning LLMs with modality-specific encoders~\citep{llasa}, or pre-training large-scale sensor language models~\citep{sensorlm, opentslm, medtsllm}. Additionally, agentic approaches have been proposed to decompose complex sensor analysis into modular actions or workflows~\citep{autoiot, personalhealthagent}.

However, most existing studies focus on single or a small number of sensor modalities and do not explicitly address the challenges of heterogeneous sensors. DriveAgent~\citep{driveagent} extends this direction to driving scenarios through a multi-agent framework with specialized agents for perception, reasoning, and decision making over multimodal sensor streams. In this work, we instead study general multimodal sensing, identifying fundamental challenges related to dense inputs and the absence of effective semantic-level fusion mechanisms, and propose a solution to address these limitations.

\subsection{Multi-Agent Collaboration}

Multi-agent collaboration enables complex task execution by assigning specialized roles and coordinating  interactions among multiple agents~\citep{camel, metagpt}. Prior work has demonstrated the effectiveness of multi-agent collaboration in coding~\citep{chatdev}, scientific research~\citep{aiscientist}, and tool use~\citep{autogen}. Recent works employ multi-agent debate mechanisms to improve reasoning accuracy by leveraging iterative deliberation to converge on more robust solutions~\citep{debate, mad, reconcile, cmd}. GAM-Agent~\citep{gamagent} extends this direction to multimodal visual reasoning by introducing uncertainty-aware control over multi-round debate. Nevertheless, existing frameworks are designed for text-centric or vision-language benchmarks, rather than sensor-driven settings with heterogeneous time-series modalities. In this work, we extend the scope of multi-agent collaboration to multimodal sensing by introducing a role-specialized framework tailored to heterogeneous sensor inputs.

\section{Discussion}
\label{sec:discussion}

\subsection{Practical Applications}
\label{sec:practical_applications}

We present \system{} as a practical framework that can be built on commodity devices such as smartphones, smartwatches, and earbuds. For example, \system{} can support context-aware assistance by integrating IMU from smartphones and smartwatches to infer situations (e.g., driving, being in the office, staying at home), and health monitoring by jointly interpreting ECG and PPG to provide early feedback on abnormal conditions. In these scenarios, synchronized sensor streams can be collected, segmented into recent sliding windows, and converted into standardized features using established preprocessing pipelines~\citep{wesad}. To facilitate practical adoption, Appendix~\ref{sec:appendix_prompts} provides task- and modality-agnostic prompt configurations, allowing practitioners to adapt \system{} to new applications by replacing placeholders with their own task definitions, modality descriptions, and sensor metadata.

\subsection{Uncertainty-Aware Fusion}

\system{} addresses sensor uncertainty implicitly through semantic interpretation and statistical agreement. This enables arbitration without requiring additional supervision or external reliability models. At the same time, explicit reliability-aware fusion could further improve performance by incorporating reliability signals as fusion weights. Promising directions include confidence-weighted voting, tool-augmented modality agents that estimate signal quality through data processing tools, and historical reliability modeling when data streams are available. We leave these extensions to future work and focus here on establishing a general, training- and model-free hybrid reasoning protocol under unknown uncertainty.
\section{Conclusion}
\label{sec:conclusion}

We propose a multi-agent collaboration framework for multimodal sensing that decomposes sensing tasks into modality-aware agents, enabling independent interpretation of heterogeneous sensor streams. Our hybrid fusion mechanism navigates the trade-off between semantic aggregation and statistical consensus of the modality-specific outputs. While semantic reasoning remains robust under sensor failures, the statistical anchor mitigates the knowledge bias often observed in LLM-based judges that over-prioritize specific modalities. Evaluations across five diverse sensing benchmarks confirm that \system{} outperforms the single-agent baseline by an average of $7.1\%$ in accuracy. Furthermore, by employing a single-round hybrid fusion protocol, \system{} achieves a $12.7\times$ reduction in fusion token costs while matching the performance of state-of-the-art multi-agent debate methods. In conclusion, \system{} provides a scalable, efficient, and robust solution for grounding LLMs in the physical world.

\section*{Acknowledgments}
\label{sec:acknowledgments}

This work is partially supported by the Institute of Information \& communications Technology Planning \& Evaluation (IITP) grant funded by the Korea government (MSIT) (No. RS-2025-02263169, Detection and Prediction of Emerging and Undiscovered Voice Phishing). This work was also partially supported by the National Research Foundation of Korea (NRF) grant funded by the Korea government (MSIT) (RS-2024-00337007). ※ MSIT: Ministry of Science and ICT.

\section*{Limitations}
\label{sec:limitations}

The scale of our experiments was constrained by the computational costs associated with multimodal sensing and multi-agent collaboration. In particular, processing heterogeneous sensor features through role-specified multi-agent collaboration incurs nontrivial inference overhead. To prioritize breadth of evaluation across diverse tasks, modalities, and baselines, we therefore conducted experiments on feasible subsets of each dataset rather than the full dataset. Future work should investigate the scalability of this framework to larger datasets and long-term sensing populations, as well as strategies for reducing inference overhead without sacrificing robustness.

This work is currently limited to classification tasks, as there is no established benchmark for evaluating LLM-based multimodal sensing across broader task types. We thus curated an evaluation suite by selecting datasets with objective ground-truth labels enable rigorous and reproducible accuracy-based evaluation. As a result, our evaluation does not cover human-centric reasoning or subjective judgment tasks. Extending the framework to such domains would require specialized data collection protocols and human annotations to assess the quality, coherence, and usefulness of generated reasoning.

This work focuses primarily on establishing modality-specific agents and the hybrid fusion protocol. As a result, we did not incorporate advanced prompting strategies, such as Self-Consistency~(SC) or iterative Self-Refinement~(SR), applied on top of \system{}. Similarly, while multi-agent collaboration protocols such as ReConcile could potentially improve performance by integrating confidence signals into the semantic fusion process, we prioritized isolating and validating the effectiveness of the core protocol without introducing additional components. These advanced combinations represent promising future directions for enhancing the reasoning capabilities of both modality agents and fusion agents.


We utilized non-fine-tuned LLMs to demonstrate the generalizability of our framework. We expect that further specialization, such as fine-tuning agents on modality-specific data or integrating Retrieval-Augmented Generation~(RAG) would further improve performance. By demonstrating that role specialization and structured data instructions alone yield significant gains, we establish a foundational step toward more complex multimodal systems. As sensor-specialized LLMs continue to emerge, \system{} can serve as a guideline protocol for designing effective collaboration among heterogeneous, modality-aware agents.


\section*{Ethical Considerations}
\label{sec:ethical_considerations}

\noindent \textbf{Potential Risks.} \system{} is a general multimodal sensing framework that can be applied to domains including health-related tasks. In high-stakes applications such as clinical decision support or mental health assessment, incorrect predictions or misleading model-generated reasoning may lead to inappropriate user actions. As LLMs may produce hallucinated or poorly calibrated interpretations, deploying such systems without adequate safeguards and human oversight may introduce safety risks. We emphasize that \system{} is intended for research and exploratory use only. Any real-world deployment in safety-critical contexts should incorporate rigorous validation, regulatory compliance, and human-in-the-loop supervision. Further research is required to establish reliability guarantees, calibration mechanisms, and domain-specific safeguards before applying such systems to precision-sensitive applications.

\noindent \textbf{Use of LLMs.} We used LLMs for language polishing of the manuscript and code formatting.


\bibliography{custom}

\clearpage
\appendix

\section{Dataset Details}
\label{sec:appendix_dataset}

\newcommand{\modACC}{\tcbox[modality=blue]{ACC}\hspace{0.0em}}
\newcommand{\modGYR}{\tcbox[modality=gray]{GYR}\hspace{0.0em}}
\newcommand{\modANG}{\tcbox[modality=purple]{ANG}\hspace{0.0em}}
\newcommand{\modEMG}{\tcbox[modality=orange]{EMG}\hspace{0.0em}}
\newcommand{\modEDA}{\tcbox[modality=yellow]{EDA}\hspace{0.0em}}
\newcommand{\modECG}{\tcbox[modality=red]{ECG}\hspace{0.0em}}
\newcommand{\modPPG}{\tcbox[modality=pink]{PPG}\hspace{0.0em}}
\newcommand{\modEEG}{\tcbox[modality=teal]{EEG}\hspace{0.0em}}
\newcommand{\modRESP}{\tcbox[modality=cyan]{RESP}\hspace{0.0em}}
\newcommand{\modTEMP}{\tcbox[modality=green]{TEMP}\hspace{0.0em}}
\newcommand{\modEOG}{\tcbox[modality=brown]{EOG}\hspace{0.0em}}
\newcommand{\modMAG}{\tcbox[modality=magenta]{MAG}\hspace{0.0em}}
\newcommand{\modHR}{\tcbox[modality=darkgray]{HR}\hspace{0.0em}}

\newcommand{\modChest}{\tcbox[modality=white]{Chest}\hspace{0.0em}}
\newcommand{\modWrist}{\tcbox[modality=white]{Wrist}\hspace{0.0em}}
\newcommand{\modFpzCz}{\tcbox[modality=white]{Fpz-Cz}\hspace{0.0em}}
\newcommand{\modPzOz}{\tcbox[modality=white]{Pz-Oz}\hspace{0.0em}}
\newcommand{\modLeftArm}{\tcbox[modality=white]{Left arm}\hspace{0.0em}}
\newcommand{\modRightArm}{\tcbox[modality=white]{Right arm}\hspace{0.0em}}
\newcommand{\modEar}{\tcbox[modality=white]{Ear}\hspace{0.0em}}
\newcommand{\modLeftWrist}{\tcbox[modality=white]{Left wrist}\hspace{0.0em}}
\newcommand{\modRightWrist}{\tcbox[modality=white]{Right wrist}\hspace{0.0em}}
\newcommand{\modWaist}{\tcbox[modality=white]{Waist}\hspace{0.0em}}
\newcommand{\modHand}{\tcbox[modality=white]{Hand}\hspace{0.0em}}
\newcommand{\modAnkle}{\tcbox[modality=white]{Ankle}\hspace{0.0em}}

\newcolumntype{T}{>{\arraybackslash}m{5cm}}
\newcolumntype{M}{>{\arraybackslash}m{8cm}}

\begin{table*}[ht]
\small
\centering
\begin{tabularx}{\textwidth}{LLCM}
\toprule
Dataset & Task & \#Classes & Sensor modalities \\
\midrule
WESAD & Affective state recognition & 3 & \modACC \modECG \modEMG \modEDA \modTEMP \modRESP (\modChest) \newline \modACC \modPPG \modEDA \modTEMP (\modWrist) \\ 
\midrule
SleepEDF & Sleep stage classification & 5 & \modEEG (\modFpzCz) \modEEG (\modPzOz) \modEOG \modEMG \modRESP \\
\midrule
ActionSense & Kitchen activity recognition & 5 & \modACC \modANG \modEMG (\modLeftArm), \modACC \modANG \modEMG (\modRightArm) \\
\midrule
MMFit & Gym exercise recognition & 10 & \modACC \modGYR$\times$(\modEar \modLeftWrist \modRightWrist \modWaist) \newline \modMAG (\modWaist) \modHR$\times$(\modLeftWrist \modRightWrist) \\
\midrule
PAMAP2 & Daily activity recognition & 12 & \modACC \modGYR$\times$(\modHand \modChest \modAnkle) \\
\bottomrule
\end{tabularx}
\caption{Summary of multimodal sensing datasets and their respective sensor modalities.}
\label{tab:datasets}
\end{table*}
\newcolumntype{F}{>{\arraybackslash}m{13.5cm}}
\begin{table*}[t]
\small
\centering
\begin{tabularx}{\textwidth}{CF}
\toprule
Sensor & Extracted features \\
\midrule
\modACC \modGYR & \multirow{2}{13.5cm}{Mean, std, and absolute integral for each axis ($x, y, z$) and magnitude. Peak frequency per axis.} \\
\modMAG \modANG &  \\
\midrule
\multirow{4}{*}{\modECG\modPPG} & Heart rate (HR): Mean and std of HR derived from inter-beat intervals. \\
& HR variability (HRV): RMSSD, pNN50, TINN, and std. \\
& Frequency domain: Power in ULF (0.01–0.04 Hz), LF (0.04–0.15 Hz), HF (0.15–0.4 Hz), and UHF (0.4–1.0 Hz). Total power, LF/HF ratio, relative powers, and normalized LF/HF components. \\
\midrule
\multirow{3}{*}{\modEDA} & Mean, std, min, max, slope, and dynamic range (5 Hz low-pass filtered). \\
& Tonic (SCL): Mean, std, and correlation with time. \\
& Phasic (SCR): Mean, std, event count, sum of magnitudes, total duration, and area under curve (AUC). \\
\midrule
\multirow{2}{*}{\modEMG} & Chain 1 (high-pass): Mean, std, dynamic range, absolute integral, median, 10th/90th percentiles. Mean/median/peak frequency and spectral energy across seven bands (0--350 Hz). \\
& Chain 2 (50 Hz low-pass): Peak count, mean/std/sum of peak amplitudes, and normalized sum of amplitudes. \\
\midrule
\modRESP & Inhalation/exhalation durations (mean, std, ratio), stretch, inspiration volume, respiration rate, and average cycle duration (0.1--0.35 Hz bandpass). \\
\midrule
\modTEMP & Mean, std, min, max, slope, and dynamic range. \\
\midrule
\multirow{4}{*}{\modEEG} & Frequency bands: Delta, theta, alpha, beta, spindle, K-complex, and sawtooth. \\
& Band features: Mean, std, variance, dynamic range, peak count, zero-crossing rate, variance of first-order difference, and absolute power (Welch's method). \\
& Ratios: Delta/theta, theta/alpha, alpha/beta, and (delta+theta)/(alpha+beta). \\
\midrule
\multirow{3}{*}{\modEOG} & Time Domain: Mean, std, variance, dynamic range, zero-crossings, and first-order difference variance. \\
& Eye movements: Large movement count ($>120\mu V$ within 1.5s) and difference variance (clean signal). \\
& Spectral: Slow (0.5--2 Hz) and rapid (2--5 Hz) power ratios relative to total power (0.5--30 Hz). \\
\bottomrule
\end{tabularx}
\caption{Summary of features extracted for different sensor types.}
\label{tab:features}
\end{table*}

\noindent \textbf{Overview.} Table~\ref{tab:datasets} summarizes the tasks and sensor modalities for each dataset. The datasets include five to eleven modalities per task, covering 12 distinct sensor types collected from diverse devices and body locations, demonstrating broad diversity in sensing configurations.

\noindent \textbf{Data Preprocessing.} Following the protocols established in the original publications~\citep{wesad, sleepedf, actionsense, mmfit, pamap2}, raw sensor data were segmented using sliding windows. To prevent data leakage between few-shot examples and target samples, we employed non-overlapping windows (where step size equals window size). For WESAD, we focused on three affective states: baseline, stress, and amusement. For SleepEDF, we utilized the standard sleep stages (W, N1, N2, N3, and REM), excluding infrequent classes. In ActionSense, we grouped the labels into five high-level categories (spreading, peeling/slicing, jar operations, wiping, and tableware tasks) following the categorization in \citet{actionsense}; ``pouring'' was excluded due to insufficient samples. We utilized data from both the left and right armbands. For MMFit, we excluded data from the left smartphone due to sample sparsity. For PAMAP2, we selected 12 of the 18 original activity classes (e.g., walking, cycling, ironing) that provided sufficient sample density.

\begin{table*}
\centering
\renewcommand{\arraystretch}{1.1}
\small
\begin{tabularx}{0.96\textwidth}{LD@{\hspace{6pt}}D@{\hspace{6pt}}D@{\hspace{6pt}}D@{\hspace{6pt}}D@{\hspace{6pt}}D}
\toprule
Method & WESAD & SleepEDF & ActionSense & MMFit & PAMAP2 & Avg. \\
\midrule

\rowcolor{gray!10}
\multicolumn{7}{l}{\textit{Modality agents} (single-round)} \\
\rowcolor{white}
$+$ Debate & \underline{0.907} \scalebox{0.8}{$\pm$ 0.024} & 0.588 \scalebox{0.8}{$\pm$ 0.031} & 0.607 \scalebox{0.8}{$\pm$ 0.031} & 0.953 \scalebox{0.8}{$\pm$ 0.012} & 0.551 \scalebox{0.8}{$\pm$ 0.026} & 0.721 \scalebox{0.8}{$\pm$ 0.025} \\
$+$ MAD & 0.847 \scalebox{0.8}{$\pm$ 0.030} & 0.551 \scalebox{0.8}{$\pm$ 0.030} & 0.591 \scalebox{0.8}{$\pm$ 0.030} & 0.960 \scalebox{0.8}{$\pm$ 0.011} & \textbf{0.581} \scalebox{0.8}{$\pm$ 0.026} & 0.706 \scalebox{0.8}{$\pm$ 0.025} \\
$+$ CMD & 0.899 \scalebox{0.8}{$\pm$ 0.025} & 0.575 \scalebox{0.8}{$\pm$ 0.032} & \underline{0.608} \scalebox{0.8}{$\pm$ 0.031} & \underline{0.964} \scalebox{0.8}{$\pm$ 0.011} & 0.553 \scalebox{0.8}{$\pm$ 0.026} & 0.720 \scalebox{0.8}{$\pm$ 0.025} \\
$+$ ReConcile & 0.886 \scalebox{0.8}{$\pm$ 0.026} & 0.557 \scalebox{0.8}{$\pm$ 0.030} & 0.596 \scalebox{0.8}{$\pm$ 0.031} & 0.944 \scalebox{0.8}{$\pm$ 0.013} & \underline{0.566} \scalebox{0.8}{$\pm$ 0.026} & 0.710 \scalebox{0.8}{$\pm$ 0.025} \\
$+$ Semantic Fusion & 0.825 \scalebox{0.8}{$\pm$ 0.031} & 0.580 \scalebox{0.8}{$\pm$ 0.031} & 0.605 \scalebox{0.8}{$\pm$ 0.033} & \underline{0.964} \scalebox{0.8}{$\pm$ 0.011} & 0.559 \scalebox{0.8}{$\pm$ 0.026} & 0.707 \scalebox{0.8}{$\pm$ 0.026} \\
$+$ Statistical Fusion & \textbf{0.927} \scalebox{0.8}{$\pm$ 0.021} & \underline{0.592} \scalebox{0.8}{$\pm$ 0.032} & 0.597 \scalebox{0.8}{$\pm$ 0.033} & 0.960 \scalebox{0.8}{$\pm$ 0.011} & 0.534 \scalebox{0.8}{$\pm$ 0.026} & \underline{0.722} \scalebox{0.8}{$\pm$ 0.026} \\
\textbf{\system{}} & 0.880 \scalebox{0.8}{$\pm$ 0.029} & \textbf{0.600} \scalebox{0.8}{$\pm$ 0.031} & \textbf{0.611} \scalebox{0.8}{$\pm$ 0.031} & \textbf{0.967} \scalebox{0.8}{$\pm$ 0.010} & 0.558 \scalebox{0.8}{$\pm$ 0.026} & \textbf{0.723} \scalebox{0.8}{$\pm$ 0.026} \\

\bottomrule
\end{tabularx}
\caption{Comparison of \system{} with multi-agent debate baselines under equal token cost (without iterative debate rounds). Results for the individual semantic and statistical fusion agents are also reported for reference. The best and second-best accuracies are highlighted in bold and \underline{underline}, respectively.}
\label{tab:round0_results}

\end{table*}

\noindent \textbf{Feature Extraction.} Features were extracted according to dataset-specific guidelines or established signal processing methods~\citep{wesad, sleepedf_features}, as summarized in Table~\ref{tab:features}. To evaluate system robustness against sensor failure, we simulated 10\%, 30\%, and 50\% modality dropout rates by randomly masking sensor streams with zeros prior to feature extraction. Detailed task descriptions, class definitions, and feature extraction procedures were stored as metadata and dynamically parsed into the LLM prompts to provide environmental context.

\noindent \textbf{Data Split.} We split each dataset into an example set (for 1-shot in-context learning) and a test set to prevent data leakage. We adopt a within-subject split strategy, assuming each user provides a single data instance per class with minimal effort. Sensor data such as EEG exhibit substantial inter-subject domain shift~\citep{eegdomainshift, domainshift}, which even remains an open challenge in the sensing community. Since LLMs are highly sensitive to in-context examples, cross-subject examples can incur significant performance variance. As our primary goal is to analyze \textit{multimodal fusion behavior} under controlled conditions, we adopt within-subject calibration to isolate fusion effects from confounding domain-shift factors.

\noindent \textbf{Data Subsampling.} Due to the high token cost of processing heavy sensor feature sequences with multiple LLMs (\texttt{gpt-oss-20B}, \texttt{gpt-oss-120B}, \texttt{Llama-3.1-8B-Instruct}, and \texttt{Llama-4-Scout-17B-16E}) and multi-agent baselines that require iterative debate rounds, we evaluate \system{} on balanced dataset subsets. We sample 50 instances per class for WESAD, SleepEDF, and ActionSense (datasets with $<$10 classes), and 30 instances per class for MMFit and PAMAP2 (datasets with $\geq$10 classes), resulting in test sets of 150–360 samples per task. Subsampling was performed uniformly while preserving the original class and subject distributions, and the same subsets were used across all baselines to ensure fair comparison. This strategy was adopted solely to make comparison across methods computationally feasible under identical budgets, rather than as a requirement of the proposed framework.

We construct a single maximally large and diverse test split for each dataset to maximize sample coverage under limited computational budgets. Since \system{} is training-free and deterministic (temperature$=0$ for all methods except Self-Consistency), the main source of randomness arises from dataset subsampling. We therefore report standard deviations using 1,000-iteration bootstrap resampling to estimate variance over samples. This evaluation protocol is adopted to prioritize broad benchmark coverage with multiple tasks and sensors, rather than to restrict statistical rigor.

\begin{table*}
\centering
\renewcommand{\arraystretch}{1.0}
\smaller
\begin{tabularx}{0.96\textwidth}{LD@{\hspace{6pt}}D@{\hspace{6pt}}D@{\hspace{6pt}}D@{\hspace{6pt}}D@{\hspace{6pt}}D}
\toprule
Method & WESAD & SleepEDF & ActionSense & MMFit & PAMAP2 & Avg. \\
\midrule
\rowcolor{gray!15}
\multicolumn{7}{l}{\textit{Single-agent} baselines} \\
\rowcolor{white}
Single Agent & 0.787 \scalebox{0.8}{$\pm$ 0.034} & 0.498 \scalebox{0.8}{$\pm$ 0.033} & 0.569 \scalebox{0.8}{$\pm$ 0.032} & 0.772 \scalebox{0.8}{$\pm$ 0.043} & 0.493 \scalebox{0.8}{$\pm$ 0.019} & 0.624 \scalebox{0.8}{$\pm$ 0.015} \\
$+$ Self-Consistency & 0.779 \scalebox{0.8}{$\pm$ 0.034} & 0.513 \scalebox{0.8}{$\pm$ 0.032} & 0.537 \scalebox{0.8}{$\pm$ 0.031} & 0.859 \scalebox{0.8}{$\pm$ 0.021} & 0.491 \scalebox{0.8}{$\pm$ 0.019} & 0.636 \scalebox{0.8}{$\pm$ 0.013} \\
$+$ Self-Refine & 0.617 \scalebox{0.8}{$\pm$ 0.095} & 0.536 \scalebox{0.8}{$\pm$ 0.032} & 0.501 \scalebox{0.8}{$\pm$ 0.055} & 0.734 \scalebox{0.8}{$\pm$ 0.052} & 0.492 \scalebox{0.8}{$\pm$ 0.018} & 0.576 \scalebox{0.8}{$\pm$ 0.025} \\
\midrule
\rowcolor{gray!15}
\multicolumn{7}{l}{\textit{Modality agents} $+$ multi-agent debate baselines} \\
\rowcolor{white}
$+$ Debate & 0.904 \scalebox{0.8}{$\pm$ 0.025} & 0.569 \scalebox{0.8}{$\pm$ 0.031} & 0.609 \scalebox{0.8}{$\pm$ 0.031} & 0.956 \scalebox{0.8}{$\pm$ 0.012} & 0.494 \scalebox{0.8}{$\pm$ 0.018} & 0.706 \scalebox{0.8}{$\pm$ 0.011} \\
$+$ MAD & 0.843 \scalebox{0.8}{$\pm$ 0.031} & 0.532 \scalebox{0.8}{$\pm$ 0.030} & 0.592 \scalebox{0.8}{$\pm$ 0.031} & 0.960 \scalebox{0.8}{$\pm$ 0.011} & 0.509 \scalebox{0.8}{$\pm$ 0.018} & 0.687 \scalebox{0.8}{$\pm$ 0.011} \\
$+$ CMD & 0.896 \scalebox{0.8}{$\pm$ 0.025} & 0.553 \scalebox{0.8}{$\pm$ 0.033} & 0.607 \scalebox{0.8}{$\pm$ 0.031} & 0.963 \scalebox{0.8}{$\pm$ 0.011} & 0.493 \scalebox{0.8}{$\pm$ 0.018} & 0.703 \scalebox{0.8}{$\pm$ 0.011} \\
$+$ ReConcile & 0.884 \scalebox{0.8}{$\pm$ 0.027} & 0.535 \scalebox{0.8}{$\pm$ 0.031} & 0.595 \scalebox{0.8}{$\pm$ 0.031} & 0.942 \scalebox{0.8}{$\pm$ 0.014} & 0.507 \scalebox{0.8}{$\pm$ 0.018} & 0.692 \scalebox{0.8}{$\pm$ 0.011} \\
\midrule
\rowcolor{gray!15}
\multicolumn{7}{l}{\textit{Modality agents} $+$ fusion agents (ours)} \\
\rowcolor{white}
$+$ Semantic Fusion & 0.821 \scalebox{0.8}{$\pm$ 0.032} & 0.553 \scalebox{0.8}{$\pm$ 0.031} & 0.600 \scalebox{0.8}{$\pm$ 0.031} & 0.963 \scalebox{0.8}{$\pm$ 0.011} & 0.482 \scalebox{0.8}{$\pm$ 0.018} & 0.684 \scalebox{0.8}{$\pm$ 0.012} \\
$+$ Statistical Fusion & 0.925 \scalebox{0.8}{$\pm$ 0.022} & 0.573 \scalebox{0.8}{$\pm$ 0.031} & 0.598 \scalebox{0.8}{$\pm$ 0.030} & 0.959 \scalebox{0.8}{$\pm$ 0.011} & 0.475 \scalebox{0.8}{$\pm$ 0.018} & 0.706 \scalebox{0.8}{$\pm$ 0.011} \\
\textbf{\system{}} & 0.877 \scalebox{0.8}{$\pm$ 0.027} & 0.581 \scalebox{0.8}{$\pm$ 0.031} & 0.611 \scalebox{0.8}{$\pm$ 0.031} & 0.966 \scalebox{0.8}{$\pm$ 0.011} & 0.488 \scalebox{0.8}{$\pm$ 0.018} & 0.705 \scalebox{0.8}{$\pm$ 0.011} \\
\bottomrule
\end{tabularx}
\caption{Comparison of \system{} with baseline methods in terms of macro-F1.}
\label{tab:main_f1_results}
\end{table*}

\section{Results in Macro-F1}
\label{sec:appendix_f1score}

Table~\ref{tab:main_f1_results} reports the main results in terms of macro-F1. Since the evaluated datasets are class-balanced, the macro-F1 results show trends similar to the accuracy results in Table~\ref{tab:main_results}.

\section{Multi-Agent Debate Performance without Iterative Rounds}
\label{sec:appendix_round0}

In our main experiments, multi-agent debate baselines (Debate, MAD, CMD, ReConcile) employ two additional iterative rounds to refine initial predictions. These additional rounds incur substantially higher token cost than \system{}. Without these iterative rounds, Debate and CMD reduce to majority voting, equivalent to our statistical fusion agent, while MAD reduces to an LLM judge, equivalent to our semantic fusion agent. ReConcile differs by applying confidence-weighted voting.

To compare performance under comparable token budgets with \system{}, we evaluate all baselines without iterative debate rounds. Table~\ref{tab:round0_results} reports the results. Debate and CMD exhibit identical performance to statistical fusion, and MAD mirrors semantic fusion, as expected. ReConcile shows reduced accuracy (71.0\%) when iterative refinement is removed, performing $1.3\%$ lower than \system{}. Overall, \system{} consistently outperforms debate-based baselines under equal token cost, demonstrating that its hybrid fusion achieves superior accuracy without relying on expensive iterative deliberation.

\section{Comparison to Supervised Models}
\label{app:supervised_fusion}

While \system{} is designed as a training-free framework, we compare it to a supervised baseline to provide an upper-bound reference. We evaluate a Linear Discriminant Analysis (LDA) classifier, following prior multimodal fusion work~\citep{wesad}, using the same feature representations provided to the LLM prompts. For each dataset, we train LDA on 60\% of the samples without overlap with the test set. Table~\ref{tab:supervised_vs_consensus} reports the results.

\begin{table}[t]
\centering
\renewcommand{\arraystretch}{1.0}
\begin{tabular}{lcc}
\toprule
Dataset & Supervised & \system{} \\
\midrule
WESAD       & 0.903 \scalebox{0.8}{$\pm$ 0.032} & 0.880 \scalebox{0.8}{$\pm$ 0.029} \\
SleepEDF    & 0.829 \scalebox{0.8}{$\pm$ 0.004} & 0.600 \scalebox{0.8}{$\pm$ 0.031} \\
ActionSense & 0.917 \scalebox{0.8}{$\pm$ 0.013} & 0.611 \scalebox{0.8}{$\pm$ 0.031} \\
MMFit       & 0.999 \scalebox{0.8}{$\pm$ 0.004} & 0.967 \scalebox{0.8}{$\pm$ 0.010} \\
PAMAP2      & 0.944 \scalebox{0.8}{$\pm$ 0.010} & 0.558 \scalebox{0.8}{$\pm$ 0.026} \\
Avg.        & 0.918 \scalebox{0.8}{$\pm$ 0.013} & 0.723 \scalebox{0.8}{$\pm$ 0.026} \\
\bottomrule
\end{tabular}
\caption{Comparison between a supervised model and \system{} in terms of accuracy. The supervised model is trained on the same feature representations, while \system{} operates in a training-free setting.}
\label{tab:supervised_vs_consensus}
\end{table}

The supervised baseline achieves strong performance by leveraging labeled training data. Although \system{} operates without task-specific training or retraining, it still performs strongly on selected datasets, including WESAD and MMFit. This suggests that \system{} can be useful in deployment settings where labeled data are limited or training is infeasible.

\begin{table*}
\centering
\renewcommand{\arraystretch}{1.0}
\smaller
\begin{tabularx}{0.96\textwidth}{LD@{\hspace{6pt}}D@{\hspace{6pt}}D@{\hspace{6pt}}D@{\hspace{6pt}}D@{\hspace{6pt}}D}
\toprule
Method & WESAD & SleepEDF & ActionSense & MMFit & PAMAP2 & Avg. \\
\midrule
Single-Agent & 0.780 \scalebox{0.8}{$\pm$ 0.033} & 0.512 \scalebox{0.8}{$\pm$ 0.033} & 0.544 \scalebox{0.8}{$\pm$ 0.031} & 0.907 \scalebox{0.8}{$\pm$ 0.017} & 0.578 \scalebox{0.8}{$\pm$ 0.026} & 0.664 \scalebox{0.8}{$\pm$ 0.028} \\
Semantic Fusion & 0.833 \scalebox{0.8}{$\pm$ 0.030} & 0.536 \scalebox{0.8}{$\pm$ 0.032} & 0.556 \scalebox{0.8}{$\pm$ 0.031} & 0.963 \scalebox{0.8}{$\pm$ 0.011} & 0.564 \scalebox{0.8}{$\pm$ 0.026} & 0.691 \scalebox{0.8}{$\pm$ 0.026} \\
Statistical Fusion & 0.913 \scalebox{0.8}{$\pm$ 0.023} & 0.528 \scalebox{0.8}{$\pm$ 0.032} & 0.616 \scalebox{0.8}{$\pm$ 0.031} & 0.980 \scalebox{0.8}{$\pm$ 0.008} & 0.542 \scalebox{0.8}{$\pm$ 0.025} & 0.716 \scalebox{0.8}{$\pm$ 0.024} \\
\system{} & 0.860 \scalebox{0.8}{$\pm$ 0.028} & 0.528 \scalebox{0.8}{$\pm$ 0.032} & 0.580 \scalebox{0.8}{$\pm$ 0.031} & 0.967 \scalebox{0.8}{$\pm$ 0.010} & 0.561 \scalebox{0.8}{$\pm$ 0.026} & 0.699 \scalebox{0.8}{$\pm$ 0.025} \\
\bottomrule
\end{tabularx}
\caption{Performance under additive white Gaussian noise (AWGN) at 10 dB SNR. Results show that \system{} preserves similar relative performance trends to the clean setting under moderate sensor corruption.}
\label{tab:noise_results}
\end{table*}

\section{Robustness to Noise}
\label{app:noise_robustness}

To assess robustness under sensor corruption that may arise in practical deployments, we evaluate \system{} under Additive White Gaussian Noise (AWGN) at 10 dB SNR. Table~\ref{tab:noise_results} reports the results. Overall, the performance trends remain similar to the main results in Table~\ref{tab:main_results}, with limited degradation under additional noise. We interpret this robustness as stemming from two factors: first, the feature-based prompting relies on statistical and frequency-domain descriptors that remain informative under moderate perturbation; second, our hybrid fusion mechanism mitigates the impact of noisy modalities by aggregating complementary evidence across agents. These results support the practical applicability of \system{} in noisy environments.

\section{Prompt Templates}
\label{sec:appendix_prompts}

We provide the detailed prompts used in the single-agent baseline and \system{} for the SleepEDF~\citep{sleepedf} dataset. We present the \texttt{SYSTEM} and \texttt{USER} prompts for each specialized agent. To maintain brevity, recurring content or dataset-specific parameters are represented as placeholders (e.g., \texttt{<classes>}).

\noindent \textbf{Single-Agent Baseline.} The system prompt assigns the agent's role and details the task context. The user prompt provides a one-shot example and the multimodal features. It mandates a structured JSON output to facilitate automated parsing of the agent's reasoning.

\begin{promptbox}{\texttt{System prompt}}
You are multimodal sensing agent that solves a sensing task.
You have the following information about the task: \\
\textbf{Task}: Classify the user's sleep stage: \texttt{<classes>}, based on physiological signals collected from wearable sensors. \\
\textbf{Classes}: \texttt{<description of the classes>} \\
You will receive sensor features from multiple modalities, and you have the following information about the modality:
\{\texttt{<modality 1>}: \{"Data collection": \texttt{<data collection protocol>}, "Feature extraction": \texttt{<feature extraction methods>}\}, \texttt{<modality 2>}: ..., \textit{(repeated for all modalities)}\} \\ \\
Your goal is to analyze the features and provide a reasoned answer using your knowledge.
\end{promptbox}

\begin{promptbox}{\texttt{User prompt}}
You have received sensor features from multiple modalities: \\
\textbf{Examples:} \\
Sensor values might not always align with your inherent knowledge due to differences in data collection or processing. So, we included a few labeled examples to help your interpretation: \\
\textit{Example of} \texttt{<class 1>}: \\
  - \texttt{<feature name>}: \texttt{<value>} \\
  - \textit{(repeated for all modality features)} \\
\textit{Example of} \texttt{<class 2>}: ... \\
\textit{(repeated for all examples)} \\
\textbf{Current sample features}: ... \\ \\
Please provide your answer for the task among \texttt{<classes>} and the reasoning for your answer.
Note that the sensor features might be wrong due to the data collection or processing.
You can evaluate the quality of the features by checking the examples you have. \\ \\
Respond in the following strict JSON format: \{"REASON": "<Reasoning for the answer>", "ANSWER": "<Answer among \texttt{<classes>}>"\} \\
Do not include any additional text outside of the JSON.
\end{promptbox}

\noindent \textbf{Modality Agents.} We provide a modality agent prompt for \texttt{EEG-Pz-Oz} in the SleepEDF dataset. This agent has a modality-specific role while using the same task template as the single-agent baseline. It follows the same user prompt structure as the single-agent baseline; the only difference is that it includes only the \texttt{EEG-Pz-Oz} features instead of the full multimodal feature set.

\begin{promptbox}{\texttt{System prompt}}
You are EEG-Pz-Oz agent that solves a sensing task.\\
\texttt{<task description prompt>}
\end{promptbox}

\begin{promptbox}{\texttt{User prompt}}
You have received sensor features from EEG-Pz-Oz modality: \\
\texttt{<modality-specific example prompt>} \\
\texttt{<modality-specific sample prompt>} \\ \\
\texttt{<instruction prompt>} \\
\texttt{<formatting prompt>}
\end{promptbox}

\noindent \textbf{Semantic Fusion Agent.} We assign the role of a fusion agent and instruct it to aggregate multiple outputs from the modality agents to solve the task. The outputs from all modality agents are merged and passed to the prompt. It is then instructed to generate a final output based on its own knowledge and expertise.

\begin{promptbox}{\texttt{System prompt}}
You are a fusion agent that solves a multimodal sensing task based on interpretations from multiple sensing agents. \\
\texttt{<task description prompt>} \\ \\
You will receive reasonings and answers from multiple agents based on their interpretations of different modalities.
Your goal is to provide a final reasoned answer for the task.
\end{promptbox}

\begin{promptbox}{\texttt{User prompt}}
You have received responses from multiple sensing agents:\\
\{"EEG-Fpz-Cz": \texttt{<modality agent output>}, "EMG-submental": ..., \textit{(repeated for all modality agents)}\} \\ \\
Using your own knowledge and expertise, analyze the reasonings and answers and provide a final reasoned answer. \\
\texttt{<formatting prompt>}
\end{promptbox}

\noindent \textbf{Statistical Fusion Agent.} The statistical fusion agent shares the same system prompt as the semantic fusion agent but utilizes a distinct user prompt. It is provided with the \texttt{<majority-voted answer>} and tasked with justifying this consensus by identifying potential failure modes in the dissenting modality agents.

\begin{promptbox}{\texttt{User prompt}}
You have received responses from multiple sensing agents:\\
\texttt{<responses from modality agents>} \\ \\
You are on the side that the correct answer is \texttt{<majority-voted answer>} which is the majority answer.
Based on the given reasonings and answers, provide a final reasoned answer for the task. \\ \\
For agents who provided different answers from \texttt{<majority-voted answer>}, explain why their reasoning is likely affected by noise or unreliable signal interpretation. 
Stay consistent with the position that the correct answer is likely \texttt{<majority-voted answer>}. \\
\texttt{<formatting prompt>}
\end{promptbox}

\noindent \textbf{Hybrid Fusion Agent.} The hybrid fusion agent is guided by a system prompt that assigns it the role of a coordinator for reconciling the outputs of the fusion agents. It is provided with (i) the modality agents' outputs and (ii) the responses from both semantic and statistical fusion agents. Its task is to evaluate these responses and provide the final consolidated output.

\begin{promptbox}{\texttt{System prompt}}
You are a coordinator agent that solves a multimodal sensing task based on interpretations from multiple sensing agents. \\
\texttt{<task description prompt>} \\ \\
You will receive reasonings and answers from multiple agents, each interpreting different modalities, as well as two fusion agents that have already aggregated these responses in different ways.
Your goal is to coordinate the responses from the fusion agents and provide a final reasoned answer for the task.
\end{promptbox}

\begin{promptbox}{\texttt{User prompt}}
You have received the following information. \\
1. Original responses from individual modality agents: \\
\texttt{<responses from modality agents>} \\ \\
2. Responses from fusion agents: \\
\texttt{<response from the semantic fusion agent>} \\
\texttt{<response from the statistical fusion agent>} \\ \\
Your task is to determine the responses from the fusion agents and provide a final reasoned answer for the task. \\
\texttt{<formatting prompt>}
\end{promptbox}

\section{Qualitative Analysis}
\label{sec:appendix_outputs}

We present outputs from the single-agent baseline and \system{} to illustrate the collaborative decision-making process, using examples from the SleepEDF dataset.

\noindent \textbf{Single-Agent Baseline.} In this example, the ground-truth label is REM, but the single-agent baseline incorrectly predicts N2. While the agent performs plausible reasoning on the spectral features, it fails to integrate the subtle respiratory cues that output REM from light sleep.

\begin{promptbox}{\texttt{LLM response} (Sinlge-agent)}
\{"REASON": "The ratio of (delta+theta) to (alpha+beta) for both Fpz-Cz (9.34) and Pz-Oz (3.75) falls between the typical values for wakefulness ($\approx6.1$ and 1.5) and stage N2 ($\approx10.6$ and $4.5$). The delta power (86.75) is higher than in N1 but much lower than in deep sleep (N3). Alpha and beta powers are lower than in N1/N2 but higher than in wakefulness, indicating a transition toward sleep. The k‑complex power and EMG activity are low, suggesting a lighter sleep stage. EOG shows many large eye movements and a high slow‑movement ratio, which is more consistent with wake or N1 rather than REM. Overall, the feature pattern best matches stage N2, the intermediate sleep stage.", "ANSWER": "N2"\}
\end{promptbox}

However, when isolating the respiration data via a specialized modality agent, the correct stage is identified:

\begin{promptbox}{\texttt{LLM response} (Respiration modality agent)}
\{"REASON": "The current sample shows an inhale duration longer than exhale (ratio 1.39) and an inhale mean of 2.49 s versus exhale mean of 1.79 s, which aligns with the REM example where the inhale is longer than the exhale (ratio 1.33). Other REM characteristics—such as a relatively high respiration rate ($\sim$13 breaths/min), moderate duration ($\sim$4.6 s), and moderate stretch—are also reflected in the current data. In contrast, the other stages (W, N1, N2, N3) all have inhale/exhale ratios below 1 and different mean durations. Therefore, the most consistent match is REM.", "ANSWER": "REM"\}
\end{promptbox}

\noindent \textbf{Semantic Fusion Agent.} In the previous example, while several modality agents provides incorrect classifications, the EEG-Pz-Oz and respiration agents correctly identify the REM stage. The semantic fusion agent successfully resolves this cross-modal conflict by prioritizing the high-fidelity evidence correct modality agents.

\begin{promptbox}{\texttt{LLM response} (Semantic fusion agent)}
\{"REASON": "The most reliable indicator of sleep stage is the EEG. While the frontal channel (Fpz-Cz) shows a high (delta+theta)/(alpha+beta) ratio typical of N2, the posterior channel (Pz-Oz) displays a ratio (3.75) that lies between the typical REM value ($\approx$2.6) and the N2 value ($\approx$4.5), and its other features (lower alpha/beta, lower variance) are more consistent with REM than with deeper stages. The respiratory pattern shows an inhale/exhale ratio $>$1 and a respiration rate ($\sim$13 breaths/min) characteristic of REM. EMG amplitude is low compared with wakefulness, which aligns with REM muscle atonia, although it is not as low as would be seen in deep sleep. EOG indicates a high slow‑movement ratio, but this can be influenced by sensor noise and is not a definitive marker for N3. Considering the stronger evidence from EEG (posterior channel) and respiration, the overall pattern points to REM sleep. Therefore the most plausible classification is REM.", "ANSWER": "REM"\}
\end{promptbox}

However, the semantic fusion agent is susceptible to prior-knowledge bias, where it may over-rely on a single modality output even when it is incorrect. For example, in an N1 stage sample, we observed a case where the respiration agent incorrectly suggested N2; despite the majority of other agents correctly identifying N1, the semantic agent erroneously followed the respiration agent.

\begin{promptbox}{\texttt{LLM response} (Semantic fusion agent)}
\{"REASON": "The most reliable signals for sleep staging are the EEG and respiration measures. Both EEG channels report low-amplitude, low-power activity, which is inconsistent with deep sleep (N3) that would show high-amplitude slow waves, and also inconsistent with REM, which would typically show characteristic sawtooth activity and higher dynamic range. The EOG channel's lack of large eye movements rules out REM and is compatible with both light and intermediate stages, but does not distinguish N1 from N2. The respiration data shows a moderate stretch ($\approx 2.2 \times 10^9$L), a respiration rate of 13.4 bpm, and exhale/inhale durations around 2.2 s---values that fall squarely within the typical range for stage N2 and are less compatible with the lower volume, longer duration pattern of N3 or the shorter, higher-rate pattern of REM. The EMG signal is intermediate between wakefulness and deeper stages, again pointing to a transitional stage rather than deep sleep. Given that the majority of modalities (EEG, respiration, EMG) lean toward an intermediate sleep stage and the EOG does not provide evidence for REM, the most consistent classification is stage N2.", "ANSWER": "N2"\}
\end{promptbox}

\noindent \textbf{Statistical Fusion Agent.} To mitigate semantic errors, the statistical fusion agent provides a complementary perspective by anchoring its reasoning to the majority consensus. In the aforementioned failure case, it identifies that the majority of modality agents point to N1 and provides a justification for why dissenting agents are likely erroneous.

\begin{promptbox}{\texttt{LLM response} (Statistical fusion agent)}
\{"REASON": "The two modalities that are most directly linked to sleep stage classification—central EEG (Fpz‑Cz) and submental EMG—both point to light sleep (N1). The Fpz‑Cz feature vector shows low‑amplitude, low‑power characteristics (high variance relative to N1, low delta power, low delta/theta ratio, and an alpha/beta ratio close to the N1 mean) that match the prototypical N1 profile far better than any other stage. The submental EMG is in the intermediate activity range between wakefulness and deeper stages, with mean and standard deviation values, percentiles, and peak statistics that align with published N1 values. These two signals are the gold standards for distinguishing wake, N1, N2, N3, and REM, so their agreement strongly supports an N1 classification. The other modalities provide weaker or conflicting evidence. The EOG horizontal channel, while showing low rapid‑movement power and high slow‑movement power, also exhibits an unusually high dynamic range and a very low difference variance, which are typical of measurement artifacts or electrode misplacement. Because eye‑movement activity is only one component of REM and N3, relying on it alone can misclassify light sleep as deep sleep. The EEG‑Pz‑Oz channel indicates REM based on low dynamic range and sawtooth power, yet its (delta+theta)/(alpha+beta) ratio is intermediate and the alpha/beta ratio is higher than the REM prototype, suggesting noise or cross‑talk from the adjacent frontal channel. Finally, the respiratory signal suggests N2 due to moderate breath duration and volume, but respiration alone is a poor discriminator between N2 and N3 and is highly variable across individuals. Given the central EEG and EMG consensus and the known unreliability of single‑modality cues in isolation, the most reliable conclusion is that the subject is in the N1 stage.", "ANSWER": "N1"\}
\end{promptbox}

\noindent \textbf{Hybrid Fusion Agent.} The hybrid fusion agent then evaluates these competing reasoning paths. In the previous example, it follows the statistical consensus. 

\begin{promptbox}{\texttt{LLM response} (Hybrid fusion agent)}
\{"REASON": "The most reliable evidence comes from the central EEG (Fpz‑Cz) and submental EMG, both of which are the gold standards for distinguishing sleep stages. The Fpz‑Cz channel shows low‑amplitude, low‑power activity with variance and delta power values that match the prototypical N1 profile far better than any other stage. The submental EMG is in the intermediate activity range between wakefulness and deeper stages, with mean, standard deviation, and peak statistics that align closely with published N1 values. These two signals agree on light sleep (N1). Other modalities provide conflicting or weaker cues: the Pz‑Oz channel suggests REM but its intermediate (delta+theta)/(alpha+beta) ratio and higher alpha/beta ratio point to noise; the EOG horizontal channel indicates minimal rapid eye movements, which is typical of both N1 and N2 and not decisive; the respiration signal is consistent with N2 but respiration alone is a poor discriminator. Given the consensus of the most reliable modalities and the known unreliability of single‑modality cues in isolation, the most robust conclusion is that the subject is in the N1 stage.", "ANSWER":"N1"\}
\end{promptbox}

\end{document}